\documentclass{article}

\usepackage[utf8]{inputenc} % allow utf-8 input
\usepackage[T1]{fontenc}    % use 8-bit T1 fonts
\usepackage{hyperref}       % hyperlinks
\usepackage{url}            % simple URL typesetting
\usepackage{booktabs}       % professional-quality tables
\usepackage{amsfonts}       % blackboard math symbols
\usepackage{nicefrac}       % compact symbols for 1/2, etc.
\usepackage{microtype}      % microtypography
\usepackage{xcolor}         % colors

\usepackage{amsmath}
\usepackage{amssymb}
\usepackage{mathtools}
\usepackage{amsthm}

\usepackage{defs}
\usepackage{multirow}
\usepackage{hyperref}
\usepackage{graphicx}
\usepackage{booktabs}
\usepackage[export]{adjustbox}
\usepackage{xcolor}
\usepackage{color,soul}
\usepackage{xargs}
\usepackage{caption}
\usepackage{subcaption}
\usepackage{lipsum}
\usepackage{natbib}
\usepackage{placeins}
\usepackage{xspace}
\usepackage{algorithmic}

\usepackage{hyperref}

\usepackage[accepted]{template}

\renewcommand{\algorithmiccomment}[1]{#1}

\icmltitlerunning{Parallelized Spatiotemporal Binding}

\begin{document}

\twocolumn[
\icmltitle{Parallelized Spatiotemporal Binding}



\icmlsetsymbol{equal}{*}

\begin{icmlauthorlist}
\icmlauthor{Gautam Singh}{ru,nv}
\icmlauthor{Yue Wang}{nv,usc}
\icmlauthor{Jiawei Yang}{usc}
\icmlauthor{Boris Ivanovic}{nv}
\icmlauthor{Sungjin Ahn}{equal,ka}
\icmlauthor{Marco Pavone}{equal,nv,stan}
\icmlauthor{Tong Che}{nv}
\end{icmlauthorlist}

\icmlaffiliation{ru}{Rutgers University}
\icmlaffiliation{nv}{NVIDIA Research}
\icmlaffiliation{usc}{University of Southern California}
\icmlaffiliation{ka}{KAIST}
\icmlaffiliation{stan}{Stanford University}

\icmlcorrespondingauthor{Gautam Singh}{singh.gautam@rutgers.edu}
\icmlcorrespondingauthor{Tong Che}{tongc@nvidia.com}

\icmlkeywords{Machine Learning, ICML}

\vskip 0.3in
]



\printAffiliationsAndNotice{\icmlEqualContribution} % otherwise use the standard text.

\renewcommand{\thefootnote}{$\dagger$}
\begin{abstract}

While modern best practices advocate for scalable architectures that support long-range interactions, object-centric models are yet to fully embrace these architectures. In particular, existing object-centric models for handling sequential inputs, due to their reliance on RNN-based implementation, show poor stability and capacity and are slow to train on long sequences. We introduce \textit{Parallelizable Spatiotemporal Binder} or \textit{PSB}\footnote{See our project page at \href{https://parallel-st-binder.github.io}{this link}.}, the first temporally-parallelizable slot learning architecture for sequential inputs. Unlike conventional RNN-based approaches, PSB produces object-centric representations, known as slots, for all time-steps \textit{in parallel}. This is achieved by refining the initial slots across all time-steps through a fixed number of layers equipped with causal attention. By capitalizing on the parallelism induced by our architecture, the proposed model exhibits a significant boost in efficiency. In experiments, we test PSB extensively as an encoder within an auto-encoding framework paired with a wide variety of decoder options. Compared to the state-of-the-art, our architecture demonstrates stable training on longer sequences, achieves parallelization that results in a 60\% increase in training speed, and yields performance that is on par with or better on unsupervised 2D and 3D object-centric scene decomposition and understanding.
\end{abstract}
\renewcommand{\thefootnote}{\arabic{footnote}}

\section{Introduction}

A key function of the human brain is to translate the incoming stream of sensory inputs into a mental model of the world. Studies suggest that this mental model is compositional, constructed from building blocks such as objects. Furthermore, the process by which this mental model is constructed is systematic, enabling us to interpret unfamiliar environments as a composition of familiar entities \cite{fodor1988connectionism, spelke2007core, lake2017building}. This systematic compositionality is critical for building autonomous agents because it would enable them to understand, plan, and act effectively and robustly in the physical world.

Towards this goal, the most relevant field in deep learning is that of object-centric learning \cite{iodine, space, slotattention, onbinding}. Object-centric learning aims to learn priors for grouping or binding low-level and unstructured sensory activations into a collection of vectors known as \textit{slots}. Each slot captures a higher-level compositional entity such as an object. 
Broadly, this grouping or binding emerges as a result of architectural priors combined with self-supervised learning, e.g., by performing auto-encoding with specialized encoder and decoder architectures. 

A notion that is widely acknowledged, but not yet fully embraced in object-centric learning, is to maximally utilize all available data with increasingly scalable architectures capable of capturing long-range dependencies \cite{vit, transformers}. 
Video data has been widely adopted in self-supervised learning and object-centric learning since videos contain temporal information such as object motion and behavior; learning from videos has been consistently shown to provide richer representations than training on static images \cite{sqair, savi, steve, Feichtenhofer2022MaskedAA}. However, current object-centric learning methods do not utilize the full potential of sequential datasets 
because of \textit{RNN-based} modeling \cite{savi, steve, savipp}. RNNs lead to major scaling issues---training instability on longer sequences due to gradient vanishing or exploding leads to degenerated performance and an increased training time complexity linear in sequence length \cite{pascanu2013difficulty}. On the other hand, current state-of-the-art sequence models use \textit{parallelizable} architectures \cite{transformers, s4} instead of RNNs to support fast and stable training on long sequences and capture long-range temporal dependencies.

\begin{figure*}[t]
    \centering
    \includegraphics[width=0.95\textwidth]{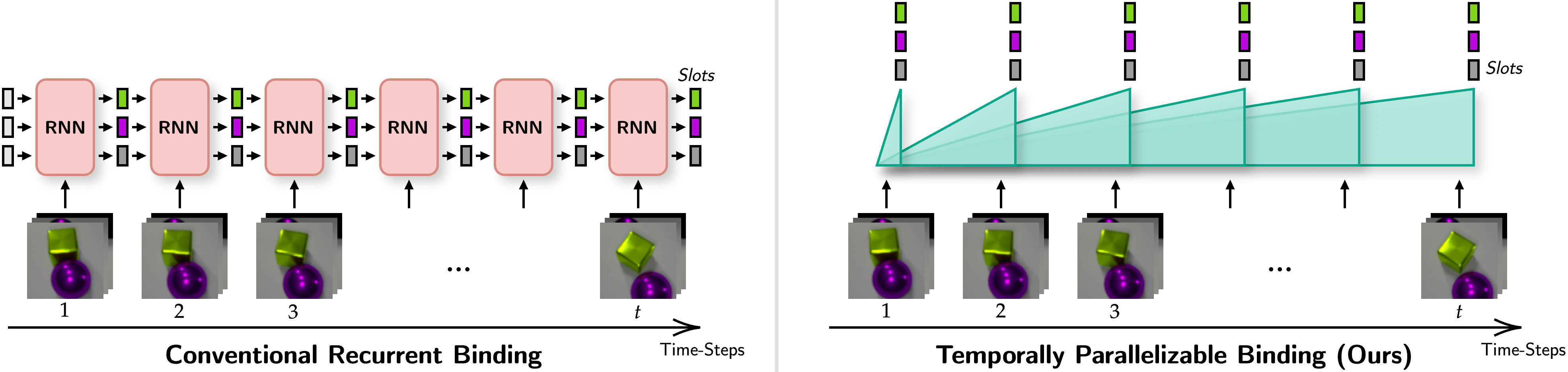}
    \caption{\textbf{Conventional Spatiotemporal Binding versus Ours.} \textit{Left:} Conventional object-centric encoders summarize sequential sensory inputs into slots via recurrence, analogous to RNNs. \textit{Right:} On the other hand, our proposed object-centric encoder achieves this without recurrence, allowing it to be parallelized over the sequence length, similarly to transformers.}
    \label{fig:one}
\end{figure*}

To address this important gap in object-centric learning, in this work, we introduce \textit{Parallelizable Spatiotemporal Binder} or PSB, the first parallelizable slot learning architecture for sequential inputs. PSB takes a sequence containing a set of input features for each time-step and generates a set of $N$-slot vectors corresponding to each time-step in a parallelizable manner. Unlike conventional object-centric learning models, which sequentially update $N$ slots through iteration over the input sequence, our novel PSB architecture eliminates the need for such sequential iteration. Instead, it produces slots for all time-steps in parallel by refining initial slots through a fixed number of layers of our proposed \textit{PSB block}. A PSB block leverages causal attention to allow each slot to directly see the input observations and the past slot states of the current layer. 
PSB is a general-purpose neural network module that can be used within any arbitrary architecture. In this work, we evaluate PSB as an encoder in various object-centric auto-encoding frameworks and demonstrate its effectiveness.

Our contributions are as follows: \textit{i)} We introduce the first temporally parallelizable object-centric binding architecture, designed to efficiently process sequential data and alleviate the common drawbacks associated with RNNs. \textit{ii)} We develop two novel parallelizable object-centric auto-encoder models by leveraging the proposed architecture as the encoder: one for 2D unposed videos and another for dynamic, posed multi-camera 3D scene videos, marking a key modeling advancement for these problems. 
\textit{iii)} Compared to the recurrent state-of-the-art baseline, our encoder shows highly stable training on longer sequences with parallelization, resulting in 1.6$\times$ faster training speed.
\textit{iv)} Our encoder, when paired with a wide variety of decoders such as an alpha-mixture decoder, an auto-regressive transformer, NeRFs, and SlotMixer, matches or exceeds the state-of-the-art recurrent baseline's performance. Specifically, with the mixture decoder for 2D videos, we observe improvements ranging from 14.7-26.8\% in FG-ARI and 2.9-7.6\% in reconstruction PSNR. For dynamic 3D scenes using NeRFs, improvements range from 7.3-121\% in slot linear-probing performance and 4-8\% in PSNR for novel view synthesis. \textit{v)} Via ablations, we obtain useful insights about the proposed design.

\section{PSB: Parallelizable Spatiotemporal Binder}

In this section, we describe our proposed architecture \textit{Parallelizable Spatiotemporal Binder} or PSB. Our architecture aims to encode or summarize a given $T$-length sequence of features $\bee_{1, 1:L}, \ldots, \bee_{T, 1:L}$ (where $\smash{\bee_{t,1:L} \in \eR^{L \times D}}$) into a $T$-length sequence of $N$ slot vectors $\bs_{1, 1:N}, \ldots, \bs_{T, 1:N}$ (where $\smash{\bs_{t, 1:N} \in \eR^{N \times D}}$). For dynamic visual scenes, the $N$ slots belonging to a specific $t$, i.e., $\bs_{t, 1:N}$, should capture an object-centric state of the world at time-step $t$. Furthermore, the $n$-th slots across all time-steps should consistently track the state of one specific object. 

Formally, PSB works by taking initial slots $\smash{\bs_{1:T, 1:N}^{(0)}}$ and transforming them conditioned on the input features $\smash{\bee_{1:T, 1:L}}$ by applying $M$ layers of our proposed PSB block: 
\begin{align*}
    \bs_{1:T, 1:N}^{(0)} &\leftarrow \text{Initialize},\\
    \bs_{1:T, 1:N}^{(i)} &\leftarrow \text{PSBBlock}_i(\bs_{1:T, 1:N}^{(i-1)}, \bee_{1:T, 1:L}), 
\end{align*}
for $i = 1, \ldots, M$. The initialization of slots can be done via learned parameters or by sampling them randomly from a learned Gaussian. The output $\smash{\bs_{1:T, 1:N}^{(M)}}$ of the last PSB block is considered as the slot representation for downstream use.

\subsection{PSB Block}
\begin{algorithm*}[t!]
\caption{\textbf{Parallelizable Spatiotemporal Binder (PSB) Block.} The algorithm receives \textit{i)} a $T$-length sequence of $N$ slot vectors denoted as $\bs_{1:T, 1:N} \in \mathbb{R}^{T\times N \times D}$, \textit{ii)} a $T$-length sequence of $L$ input feature vectors $\bee_{1:T, 1:L} \in \mathbb{R}^{T\times L \times D}$ and \textit{iii)} an optional attention mask $\boldsymbol{\alpha}\in \{0, 1\}^{T \times T}$ to enforce causal masking. The algorithm outputs the updated slots $\bs_{1:T, 1:N}$ conditioned on the input features $\bee_{1:T, 1:L}$.}\label{algo:psb_block}
\begin{algorithmic}[1]

  \STATE \textbf{Input}: $\bs\in\mathbb{R}^{T \times N \times D}$, $\bee\in\mathbb{R}^{T \times L \times D}$, $\boldsymbol{\alpha}\in \{0, 1\}^{T \times T}$ \\[0.2em]
  \STATE \textbf{Layer params}: Cross-Attention $\texttt{CA}$; Self-Attention $\texttt{SA}_1$, $\texttt{SA}_2$; LayerNorm $\texttt{LN}$; MLP $\texttt{MLP}$\\[0.2em]
  \STATE \quad \textbf{for} $t = 1\ldots T$ \textbf{in parallel} \\[0.2em]
  \STATE \qquad $\bs_{t,1:N} \mathrel{+\!\!=} \texttt{CA}\big(\texttt{q=}\texttt{LN}\left(\bs_{t,1:N}\right), \texttt{kv=}\texttt{LN}\left(\bee_{1:T,1:L}\right), \texttt{attn\_mask=}\boldsymbol{\alpha}\big)$  \hfill\COMMENT{{\color{gray} \# Slots attend input features.}}\\[0.2em]
  \STATE \quad \textbf{for} $n = 1\ldots N$ \textbf{in parallel}\\[0.2em]
  \STATE \qquad $\bs_{1:T,n} \mathrel{+\!\!=} \texttt{SA}_1\big(\texttt{qkv=}\texttt{LN}(\bs_{1:T,n}), \texttt{attn\_mask=}\boldsymbol{\alpha}\big)$  \hfill\COMMENT{{\color{gray} \# Slots with the same index self-attend across time.}}\\[0.2em]
  \STATE \quad \textbf{for} $t = 1\ldots T$ \textbf{in parallel}\\[0.2em]
  \STATE \qquad $\bs_{t,1:N} \mathrel{+\!\!=} \texttt{SA}_2\big(\texttt{qkv=}\texttt{LN}(\bs_{t,1:N})\big)$  \hfill\COMMENT{{\color{gray} \# Slots at the same time-step self-attend.}}\\[0.2em]
  \STATE \quad $\bs_{1:T,1:N} \mathrel{+\!\!=} \texttt{MLP}\big(\texttt{LN}(\bs_{1:T,1:N})\big)$  \hfill\COMMENT{{\color{gray} \# Slots undergo an MLP.}}\\[0.2em]
  \STATE \quad \textbf{return} $\bs_{1:T, 1:N}$\label{algo_step:return}\\[0.2em]
\end{algorithmic}
\end{algorithm*}

Broadly, a PSB block works by interleaving three operations: \textit{(i)} bottom-up attention by the slots on the input features, \textit{(ii)} self-attention among the slots, and \textit{(iii)} an MLP. 

\subsubsection{Bottom-Up Attention}
In this step, the slots access the bottom-up information from the inputs. For all $t=1, \ldots, T$ in parallel, we perform:
\begin{align*}
    \bs_{t,1:N} &\mathrel{+\!\!=} \texttt{CA}\big(\texttt{q=}\texttt{LN}(\bs_{t,1:N}), \texttt{kv=}\texttt{LN}(\bee_{1:T,1:L})\big),
\end{align*}
where $\texttt{LN}$ denotes layer normalization, $\texttt{CA}$ denotes multi-headed cross-attention, argument $\texttt{q}$ denotes the query and the argument $\texttt{kv}$ denotes the key and the value which are the same in this case. In implementing this cross-attention, we make three key design choices. First, we provide an option to apply \textit{causal masking} to prevent the slots from seeing the inputs of the future time-steps. This makes our model useful as a perception module in agent-learning settings where the agent typically does not have access to future observations. Second, we employ \textit{inverted-attention and renormalization} \cite{slotattention, invertedattn} to introduce competition among slots and to help them specialize to distinct objects. Third, to incorporate invariance to translation-in-time and to help the encoder generalize to any sequence length, we recommend using \textit{relative positional bias} \cite{rpb} instead of absolute positional embedding \cite{transformers} to incorporate the temporal position information in the attention process.

\subsubsection{Slot Interaction}
Next, the slots self-attend, allowing each slot to read the other slots to \textit{(i)} facilitate efficient allocation of slot resources to distinct objects, \textit{(ii)} to help the slots align across time, and \textit{(iii)} to allow each slot to become more informative by accessing the information of other slots. We execute this self-attention via two decoupled steps:

\textbf{Time-Axis Self-Attention.} In conventional recurrent object-centric encoders \cite{scalor, gswm, savi}, each slot $\bs_{t,n}$ is informed about its previous states $\bs_{<t, n}$ through an independent RNN assigned per slot. This conforms to the physical principle that distinct objects evolve largely independently of each other over time, while the temporal states of the same object are highly correlated. Incorporating this principle, we perform self-attention between all slots along the time axis sharing the same index $n$. Specifically, for all $n=1, \ldots N$ in parallel:
\begin{align*}
    \bs_{1:T,n} \mathrel{+\!\!=} \texttt{SA}_1\big(\texttt{qkv=}\texttt{LN}(\bs_{1:T,n})\big),
\end{align*}
where $\texttt{SA}_1$ denotes multi-headed self-attention and the argument $\texttt{qkv}$ denotes the query, key, and value which are the same in this case. As before, we provide the option to apply causal masking and we recommend using relative position bias to incorporate the temporal position information.

\textbf{Object-Axis Self-Attention.} Next, we let the $N$ slots of each time-step interact, i.e., for all time-steps $t=1, \ldots, T$ in parallel:
\begin{align*}
    \bs_{t,1:N} \mathrel{+\!\!=} \texttt{SA}_2\big(\texttt{qkv=}\texttt{LN}(\bs_{t,1:N})\big),
\end{align*}
where $\texttt{SA}_2$ denotes multi-headed self-attention and the argument $\texttt{qkv}$ denotes the query, key, and value which are the same in the case of self-attention.

\subsubsection{Processing Gathered Information via MLP}
To process the information gathered via the bottom-up attention and slot interaction steps, the slots are fed to an $\texttt{MLP}$ for all $t=1, \ldots, T$ and $n=1, \ldots N$ in parallel:
\begin{align*}
    \bs_{t,n} \mathrel{+\!\!=} \texttt{MLP}\big(\texttt{LN}(\bs_{t,n})\big),
\end{align*}
The resulting slots are then passed to the downstream layers. The operation of a PSB block is also summarized in Algorithm \ref{algo:psb_block}. Note that all operations described above are performed through residual connections, making the PSB block suitable for deep stacking \cite{residual}.

\section{Object-Centric Learning via Parallelizable Spatiotemporal Binder}

In this section, we outline two application scenarios of our proposed encoder.

\subsection{2D Unposed Videos}
\label{sec:appl:2d}
In this setting, we perform unsupervised object-centric representation learning from 2D unposed videos. 
For this, we adopt the video auto-encoding framework of \cite{savi, steve}. In particular, a video contains $T$ frames $\smash{\bx_1, \ldots, \bx_T}$ where each frame is an RGB image $\smash{\bx_t \in \eR^{C \times H \times W}}$. Our goal is to encode it into slot representations $\bs_{1, 1:N}, \ldots, \bs_{T, 1:N}$, where $\smash{\bs_{t,1:N} \in \eR^{N \times D}}$ denotes a collection of $N$ slot vectors for the $t$-th time-step.

To achieve this, we first encode each frame $\bx_t$ via a CNN and flatten the resulting feature map, producing $L$ feature vectors per frame: $\bee_{t} = \text{CNN}_\phi(\bx_t) \in \eR^{L\times D}$. On these features, we apply our proposed PSB encoder to obtain the slots: $\smash{\bs_{1, 1:N}, \ldots, \bs_{T, 1:N} = \text{PSB}_\phi(\bee_{1}, \ldots, \bee_{T})}$.
This slot inference process is trained in a self-supervised manner by decoding the slots and trying to reconstruct the original video frames. We consider two decoder choices. For visually simple datasets, we test an alpha-mixture decoder which is trained with MSE reconstruction loss $\cL(\phi, \ta) = ||\bx_t - \text{Decoder}_\ta(\bs_{t, 1:N}) ||^2$ \cite{slotattention}. For visually complex datasets, we test the auto-regressive image-transformer decoder \cite{slate, steve} which is trained with cross-entropy loss to reconstruct a DVAE representation of the video frames. (See Appendix \ref{ax:autoreg_tf})

\subsection{3D Posed Multi-Camera Videos}
\label{sec:appl:3d}
In this setting, we perform unsupervised object-centric representation learning on dynamic 3D scenes. In this setup, we aim to learn slot representations for a given $T$-length sequence of posed multi-camera observations denoted by $\bX_1, \ldots, \bX_T$. Here, each $\bX_t$ consists of $K$ distinct observations or \textit{views} corresponding to the $K$ cameras in the scene i.e., $\bX_t = \bx_{t,1}, \ldots, \bx_{t,K}$. Each view $\bx_{t,k}$ belongs to $\eR^{C\times H \times W}$ where $H$ and $W$ are the image height and width, respectively and $C$ is the number of channels. Since these are posed observations, $C=9$ which includes 3 channels for RGB pixel color, 3 channels for the camera ray origin, and 3 for camera ray direction.

\textbf{Novel View Prediction.} We adopt a novel view prediction framework to train the model where a certain fraction of the input views are held out and the remaining are passed to the encoder to infer the slots. The slots are then decoded via a viewpoint-conditioned decoder that tries to predict all the available views---both the held-out and the shown ones. This is a common practice in 3D scene representation learning frameworks \cite{gqn, snp, osrt}.

\textbf{Set Latent Scene Representation (SLSR).} To encode the views, we adopt the backbone of \citet{osrt}. For each time-step, we feed the $K'$ (out of $K$) visible views to a CNN to obtain a feature map. We flatten the feature maps of each view, stack them together for all $K'$ views, and feed them to a transformer. The transformer's output is known as Set Latent Scene Representation or SLSR: 
\begin{align*}
    \tilde{\bee}_{t,k} = \text{CNN}_\phi(\bx_{t,k}) && \Longrightarrow && \bee_{t} = \text{Transformer}_\phi(\tilde{\bee}_{t,1:K'}),
\end{align*}
where $\bee_t \in \eR^{L\times D}$ is the SLSR for the time-step $t$ and is a collection of $L$ feature vectors.

\textbf{Slot Learning via PSB.} Next, we feed the SLSRs $\bee_{1}, \ldots, \bee_{T}$ to our proposed encoder PSB and obtain slots: $
    \smash{\bs_{1, 1:N}, \ldots, \bs_{T, 1:N} = \text{PSB}_\phi(\bee_{1}, \ldots, \bee_{T}).}
$
We then decode the slots to reconstruct the available pixels of both the novel and the visible views. We employ viewpoint-aware decoders that take a ray (origin and direction) as input and predict the pixel color conditioned on the slot representation. In particular, to render a ray $\br$ of a time-step $t$, we can describe the decoding process as: $\hat{\bc} = \text{Decoder}_\ta(\br, \bs_{t, 1:N})$, where $\bc \in \eR^{3}$.
To train the complete model, we minimize the MSE loss of the predicted pixel against the true color of the pixel i.e., $\cL(\phi, \ta) = || \hat{\bc} - \bc ||^2$. 

\textbf{Decoder Options.} We consider two 3D decoders: NeRF \cite{nerfs} and SlotMixer \cite{osrt}. For NeRF, we maintain an MLP $g^\text{slot}_\ta$ which, for a given 3D coordinate $\bo$ and a viewing direction $\bd$, returns a color value and a density value conditioned on a slot representation: $\mathbf{c}_n, \mathbf{\sigma}_n = g^\text{slot}_\ta(\bo, \bd, \bs_n)$. Combining the outputs for $N$ slots $\bs_1, \ldots, \bs_N$, we obtain the combined density $\sigma = \sum_{n=1}^N \sigma_n$ and color $\mathbf{c} =  \sum_{n=1}^N \bc_n \sigma_n/\sigma$.
To obtain the color of an image pixel, we shoot the corresponding ray from the camera, sample $N_\text{bins} + 1$ points along the ray, and integrate the colors along the ray as:
$\sum_{i=1}^{N_\text{bins}} T_i \alpha_i \bc_i$,
where $T_i = \prod_{j=1}^{i-1} (1 - \alpha_i)$ is the transmittance, and $\alpha_i = 1 - \exp (-\sigma_i ||\bo_{i+1} - \bo_i||_2) $ is the opacity. To capture complex static topography, we investigate incorporating a static field and a sky field following \citet{emernerf}, decoupling the static field modeling from the dynamic field modeling. For SlotMixer, we use its default implementation \cite{osrt}. For a detailed description of the decoders, see Appendix \ref{ax:3ddec:nerf} and \ref{ax:3ddec:slotmixer}.

\section{Related Work}

A large body of work has emerged on the topic of learning compositional and object-centric scene representations \citep{infogan, betavae, monet, iodine, slotattention,  nem, genesis, engelcke2021genesis, anciukevicius2020object, von2020towards, onbinding, slate, chang2022object, ego, cae, rotatingfeatures, ctcae, lsd, slotdiffusion, boqsa}. Most architectures for videos start as static scene models that are extended by applying the same model recurrently on the video frames, thus making them non-parallelizable unlike ours \cite{sqair, silot, gswm, savi, iodine, op3, savipp, steve, videosaur}. This is also true for 3D-aware object-centric learning where static scene models \cite{roots, mulmon, uorf, obsurf, castrejon2022inferno, dorsal, boqsa} are sometimes extended to handle dynamic 3D scenes via recurrence \cite{crawford2020learning, dymon}. While exceptions to this exist \cite{simone, henderson2020unsupervised, Gopalakrishnan2022UnsupervisedLO}, however, these learn global slots for the entire episode instead of per time-step slots in a parallelizable manner like ours. Orthogonal to object-centric learning, efforts to resolve concerns of scalability and parallelizability of RNNs have a long history \cite{oord2016wavenet, pixelcnn, dilatedrnn, transformers, dauphin2017language, li2018independently, s4, trinh2018learning}. However, ours is the first work that tackles the question of bringing such scalability and parallelization to object-centric learning. For an extended discussion of the related work, see Appendix \ref{ax:addl_related_work}.

\begin{figure}[t]
    \centering
    \includegraphics[width=0.98\columnwidth]{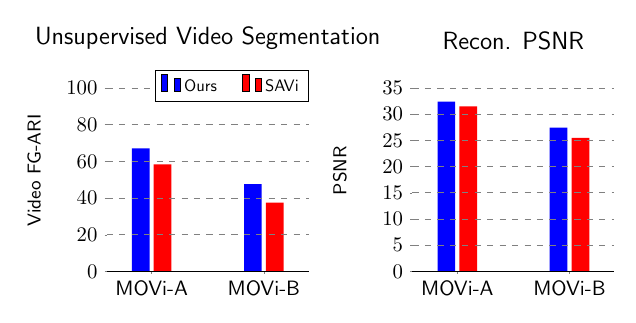}
    \includegraphics[width=0.98\columnwidth]{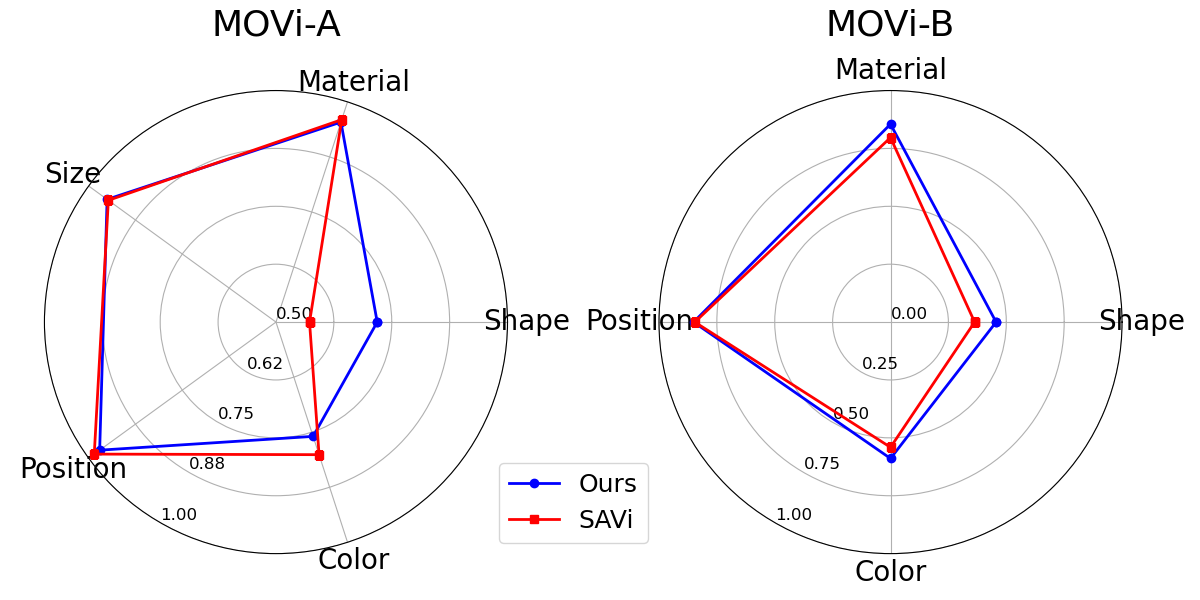}
    \vspace{-4mm}
    \caption{\textbf{Unsupervised Object-Centric Learning on MOVi-A and MOVi-B using Spatial Broadcast Decoder.} We compare our proposed encoder with the recurrence-based baseline encoder SAVi \cite{savi}. \textit{Top-Left:} Video-level FG-ARI score $(\uparrow)$. \textit{Top-Right:} Reconstruction PSNR $(\uparrow)$. \textit{Bottom:} Slot linear probing performances $(\uparrow)$. Reported are the $R^2$ score for continuous-valued object factors (position and color) and classification accuracy for categorical object factors (shape, size, and material). We observe that our encoder surpasses the recurrent baseline SAVi in terms of FG-ARI and PSNR, and does markedly better in linear-probing performance for complex factors such as the object shape.}
    \label{fig:movi-ab-ari-psnr-probes}
\end{figure}

% \clearpage
\section{Experiments}
In experiments, we demonstrate the advantages of our proposed parallelizable encoder compared to the conventional recurrent slot learning approach. We evaluate performance in two object-centric learning settings, specifically: \textit{i)} learning from unposed 2D videos, and \textit{ii)} learning from posed multi-camera videos of dynamic 3D scenes.

\subsection{Learning from 2D Unposed Videos}

\begin{figure}[t]
    \centering    
    \includegraphics[width=0.98\columnwidth]{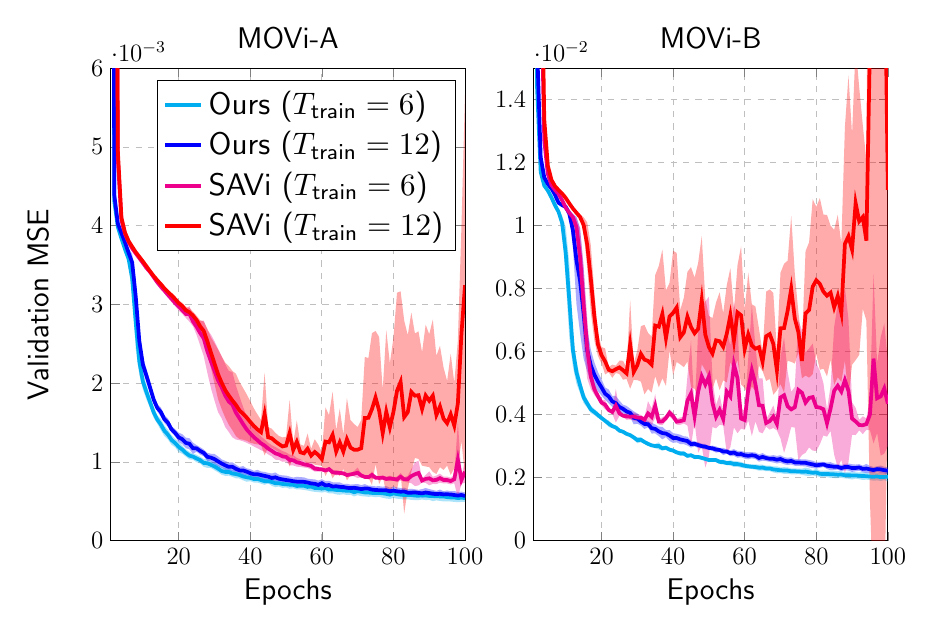}
    \includegraphics[width=0.92\columnwidth]{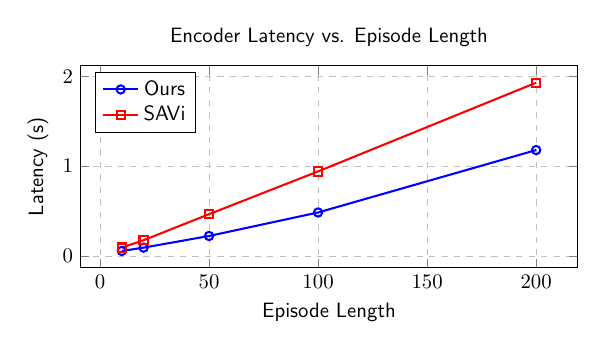}
    \vspace{-4mm}
    \caption{\textbf{Computational Drawbacks of RNN-based Object-Centric Learning.} We compare our proposed encoder with the recurrent baseline SAVi. \textit{Top:}  We show validation loss curves (mean and standard deviation computed over 5 seeds) for training runs on MOVi-A and MOVi-B. $T_\text{train}$ denotes the length of each training episode. We note that as we increase the episode length from 6 to 12, SAVi becomes highly unstable while our model continues to train smoothly. \textit{Bottom:} We report the time taken (in seconds) to perform one training step plotted as a function of the episode length. We observe a speed-up of about 1.6$\times$ over SAVi.}
    \label{fig:computation}
\end{figure}

\begin{figure}[t]
    \centering
    \includegraphics[width=1.0\columnwidth]{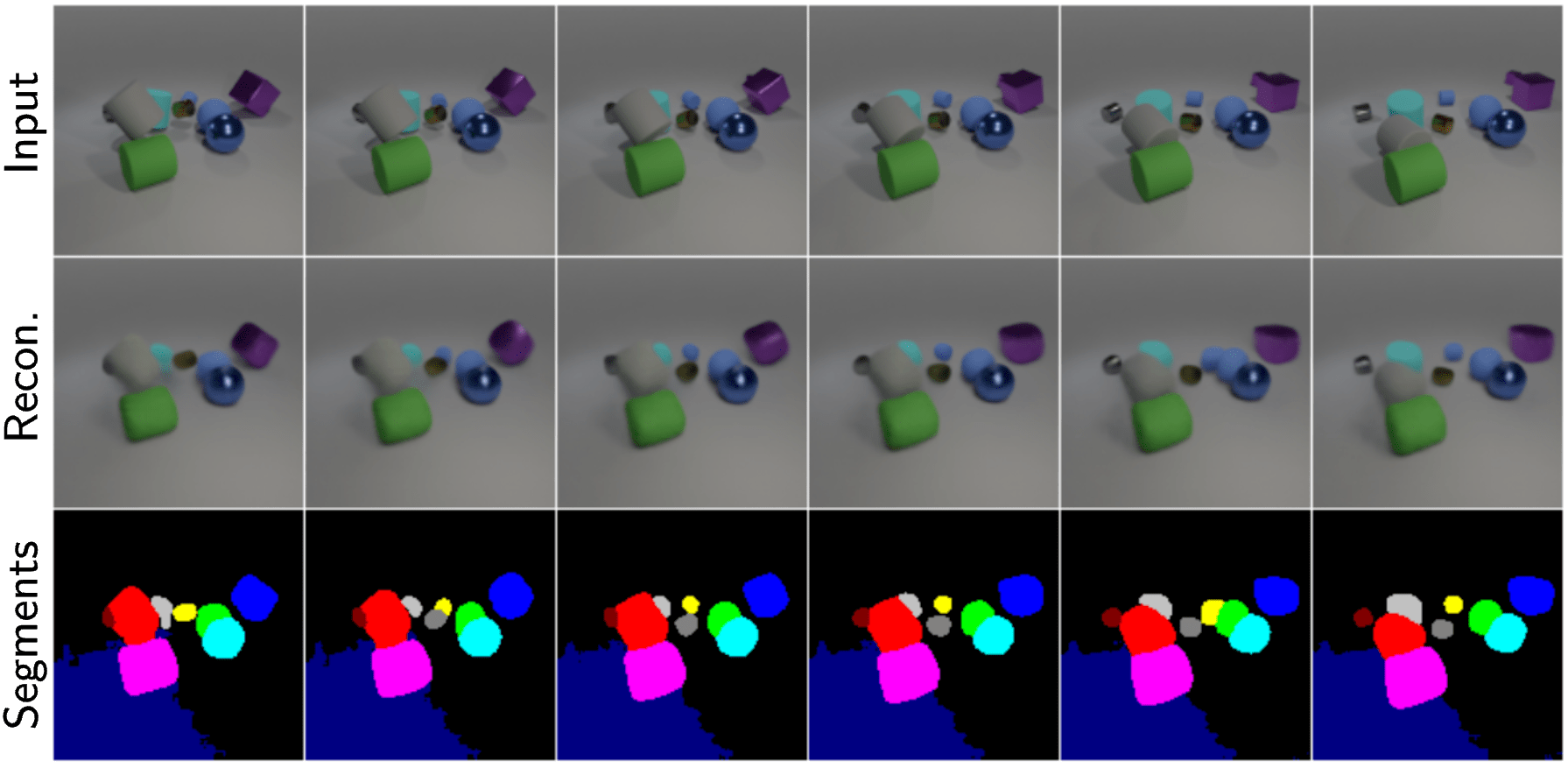}
    \vspace{-4mm}
    \caption{\textbf{Object-Centric Learning on MOVi-A using Our Proposed Encoder.} We visualize a given video and its reconstruction and decomposition into objects using the proposed model. 
    We note that object identity is consistently maintained over time as evidenced by the segment colors across frames.
    }
    \label{fig:qual-movi-a}
\end{figure}

\subsubsection{Setup}

\textbf{Datasets.} In this setting, we evaluate on the MOVi benchmark \cite{kubric} comprising five datasets: MOVi-A, MOVi-B, MOVi-C, MOVi-D, and MOVi-E. We only use the RGB frames without any auxiliary inputs or supervision. 

\textbf{Baselines.} We implement our model following the description in Section \ref{sec:appl:2d} and evaluate our proposed encoder paired with two decoders: an alpha-mixture decoder (see Appendix \ref{ax:sbd}) for visually simple MOVi-A and B datasets and an autoregressive transformer decoder (see Appendix \ref{ax:autoreg_tf}) for visually complex MOVi-C, D and E datasets. For both decoder choices, we compare the performance by substituting our proposed encoder with the current state-of-the-art and one of the most popular neural network architectures for learning sequential slot representations from videos \cite{savi, savipp}. This amounts to a comparison with unsupervised SAVi \cite{savi} when using an alpha-mixture decoder and STEVE \cite{steve}  when using an autoregressive transformer decoder.

\textbf{Metrics.} To measure the performance, we report the \textit{video-level FG-ARI} score, slot linear probing performance and the \textit{reconstruction PSNR}. The video-level FG-ARI score is a common metric \cite{savi, savipp, simone} to measure whether a video is accurately decomposed into slot-based representation and whether the decomposition is consistent across the time-steps. We take episode length to be 6. The linear probing performance for evaluating the representational informativeness of slots is measured in terms of $R^2$ score for continuous object factors and classification accuracy for the discrete object factors following \citet{dang2021evaluating} (see Appendix \ref{ax:metric:libprobe}). The reconstruction PSNR measures how well the input frames can be reconstructed from the representation. For a fair comparison, we use an identical spatial-broadcast decoder for all baselines and datasets. For MOVi-C, D, and E, where the models are trained using an autoregressive transformer decoder, we freeze the slots and train a slot-conditioned spatial-broadcast decoder for evaluating FG-ARI and PSNR.

\subsubsection{Results}

\textbf{Unsupervised Scene Decomposition and Representation.} In terms of scene decomposition performance, in Fig.~\ref{fig:movi-ab-ari-psnr-probes}, we note that our proposed encoder demonstrates a significantly superior FG-ARI score compared to the recurrent encoder baseline SAVi.
In terms of representation quality, in Fig.~\ref{fig:movi-ab-ari-psnr-probes}, we note that our encoder achieves a significantly better PSNR compared to the baseline. Furthermore, ours shows better overall linear-probing performance over the object factors, with a particularly marked improvement for the shape factor. The shape factor---being more complex than other factors like color---indicates that our encoder is capable of learning more expressive object representations than the recurrent baseline. Importantly, this improved expressiveness does not compromise the object-centric decomposition, as evidenced by our superior FG-ARI scores.

\textbf{Training Stability on Long Episodes.} Fig.~\ref{fig:computation} demonstrates the training stability of our encoder compared to the recurrent SAVi model. Owing to our encoder's ability to directly attend to any previous time-step without relying on recurrence, it exhibits high stability and trains smoothly on longer episodes. In contrast, the SAVi model displays considerable instability due to its recurrence mechanism, which is susceptible to gradient explosion.

\textbf{Training Speed on Long Episodes.} Fig.~\ref{fig:computation} also compares the duration of a single training step as a function of the episode length. Thanks to our parallelizable implementation which eliminates the need to iterate explicitly over the entire sequence length, our encoder demonstrates superior speed on very long sequences.

\begin{figure}[t]
    \centering
    \includegraphics[width=\columnwidth]{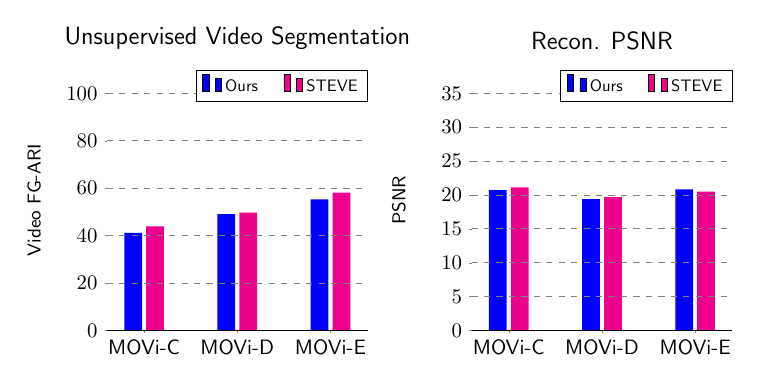}
    \vspace{-4mm}
    \caption{\textbf{Unsupervised Video Segmentation on MOVi-C, D and E using Autoregressive Image-Transformer Decoder.} We compare ours with STEVE which is based on the recurrent encoder of \citet{savi}. \textit{Left:} Video-level FG-ARI score $(\uparrow)$. \textit{Right:} Reconstruction PSNR $(\uparrow)$.}
    \label{fig:movi-cde-ari-psnr}
\end{figure}

\textbf{Compatibility with Expressive Decoders.} Expressive decoders, e.g., autoregressive image transformers, have been advantageous for object-centric learning in visually complex scenes \cite{slate, steve, lsd, slotdiffusion}. Since these decoders have been typically paired with the Slot Attention family of encoders, it becomes a question whether our proposed encoder also maintains compatibility with such powerful decoders or not. In Fig.~\ref{fig:movi-cde-ari-psnr}, we evaluate our encoder on visually complex scenes using the autoregressive transformer decoder. We find that while our segmentation performance is slightly worse, the difference is not substantial. Therefore, in scenarios where training efficiency and stability are prioritized, our proposed encoder with powerful decoders remains a preferred option.

\begin{figure}[t]
    \centering
    \begin{subfigure}[b]{0.98\columnwidth}
        \includegraphics[width=\linewidth]{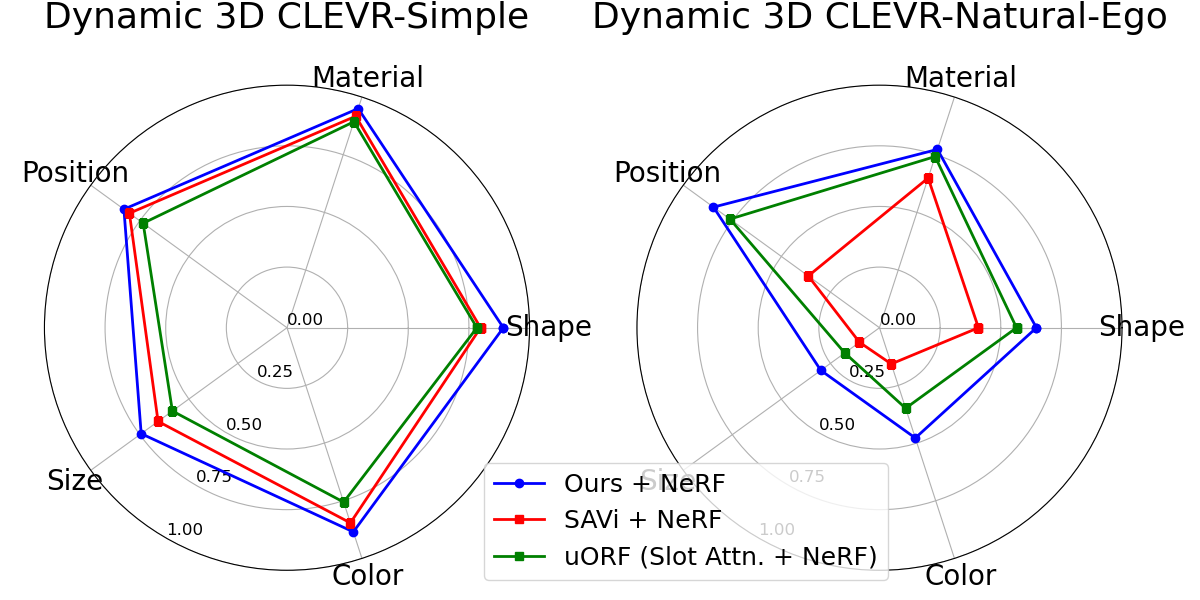}
        \caption{NeRF Decoder}
        \vspace{4mm}
        \label{fig:3d-nerf-probing}
    \end{subfigure}
    \begin{subfigure}[b]{0.98\columnwidth}
        \includegraphics[width=\linewidth]{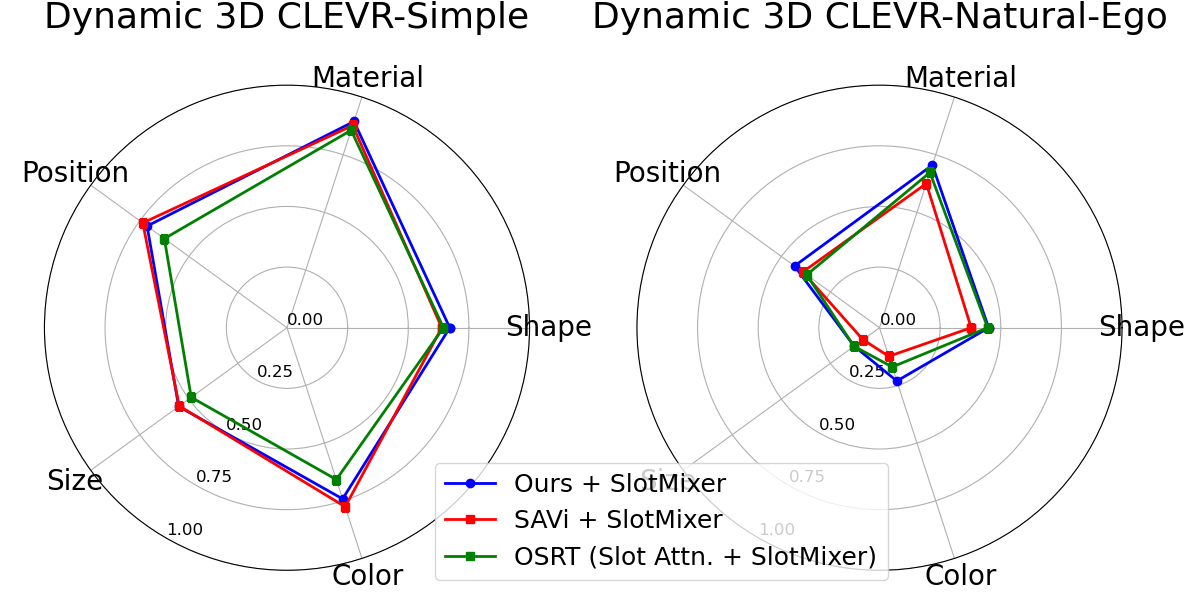}
        \caption{SlotMixer Decoder}
        \label{fig:3d-slotmixer-probing}
    \end{subfigure}
    \caption{\textbf{Comparison of Encoders in Dynamic 3D Scenes.} We compare the encoder performances trained with two decoder options: NeRF (top) and SlotMixer (bottom). 
    % We compare ours with the baselines: SAVi and per-timestep Slot Attention.
    Reported are the $R^2$ score $(\uparrow)$ for continuous-valued object factors (position, size, and color) and classification accuracy $(\uparrow)$ for categorical object factors (shape and material). With both decoders, our encoder surpasses SAVi as well as the static 3D scene models, with a noticeably large margin in the case of NeRF decoder.}
    \label{fig:combined-nerf-slotmixer}
\end{figure}
\subsection{Learning from 3D Posed Multi-Camera Videos}

\subsubsection{Setup}

\textbf{Datasets.} In this setting, we evaluate on two datasets: \textit{Dynamic 3D CLEVR-Simple} and \textit{Dynamic 3D CLEVR-Natural-Ego}. We synthesize these datasets as extensions of the CLEVR dataset to incorporate physical dynamics, multiple cameras, 3D camera pose information, moving ego observer, and visual complexity. We visualize these datasets in Fig.~\ref{fig:3d-qual} and \ref{fig:data-samples}. For more details, see Appendix \ref{ax:datasets}.

\begin{figure}
    \centering
    \begin{subfigure}[b]{1.0\columnwidth}
    \includegraphics[width=1.0\columnwidth]{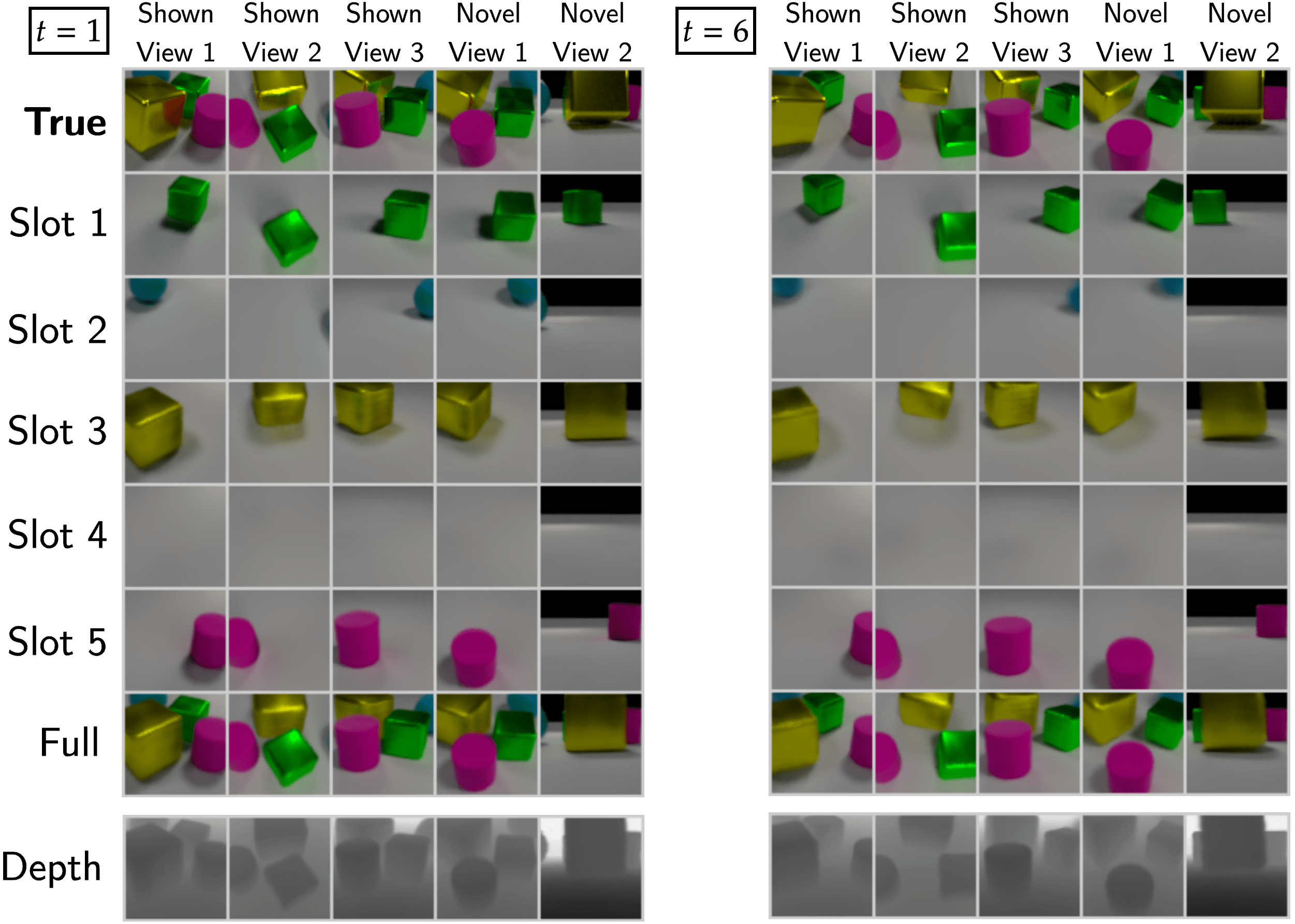}
    \caption{Dynamic 3D CLEVR-Simple}
        \vspace{2mm}
    \end{subfigure}
    \begin{subfigure}[b]{1.0\columnwidth}
        \includegraphics[width=1.0\columnwidth]{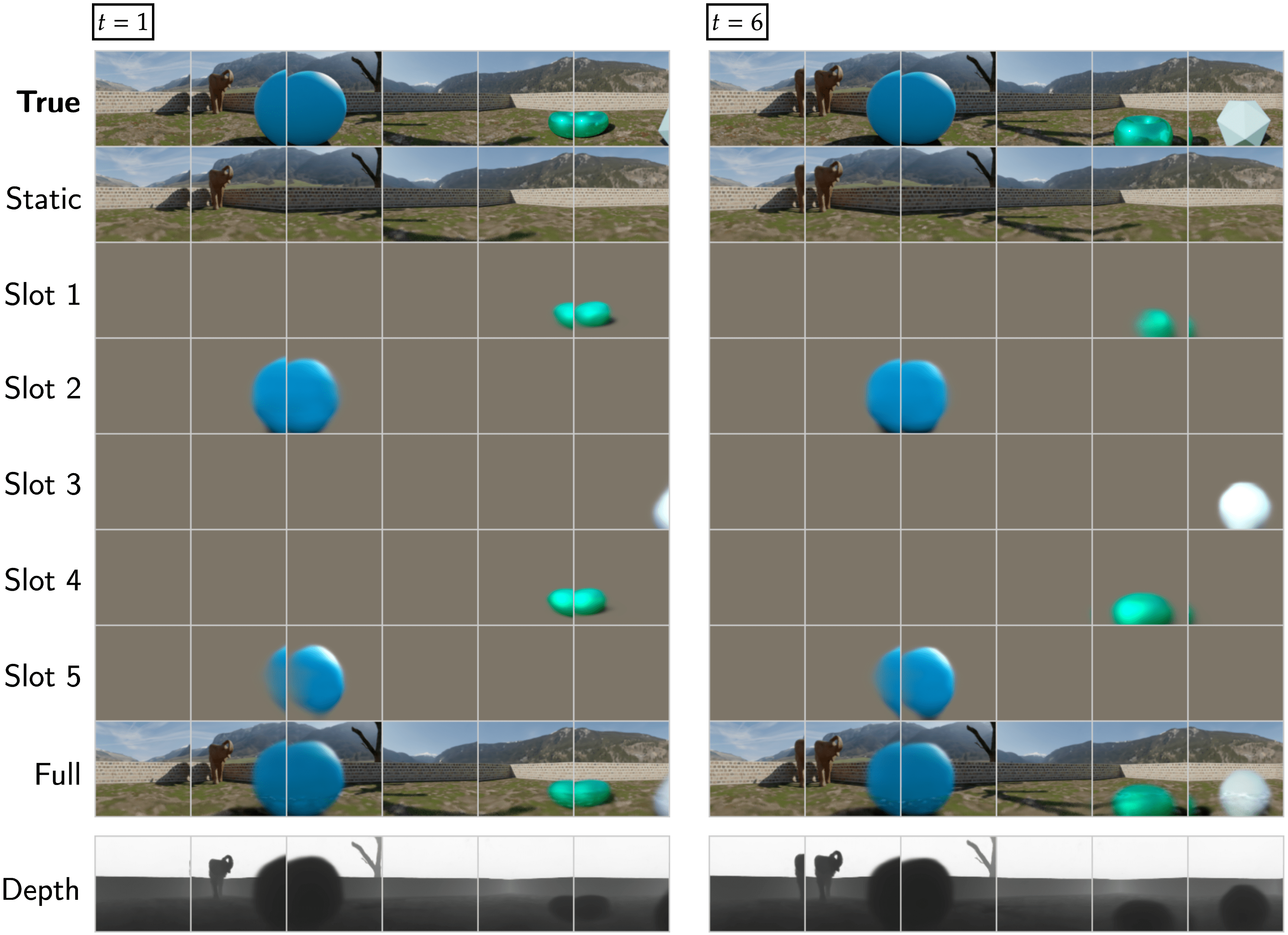}
        \caption{Dynamic 3D CLEVR-Natural-Ego}
    \end{subfigure}
    \caption{\textbf{Unsupervised Dynamic 3D Scene Understanding.} We visualize the RGB rendering of the individual slots and the RGB and depth rendering from all slots taken together for time-steps: $t=1$ and $t=6$. The model was trained as described in Section \ref{sec:appl:3d} with a NeRF decoder. We note the 3D scene decomposition, consistent alignment of slots across time-steps, and unsupervised depth inference. See the GIF version in the supplementary.}
    \label{fig:3d-qual}
\end{figure}

\begin{figure}
    \centering
    \includegraphics[width=0.96\columnwidth]{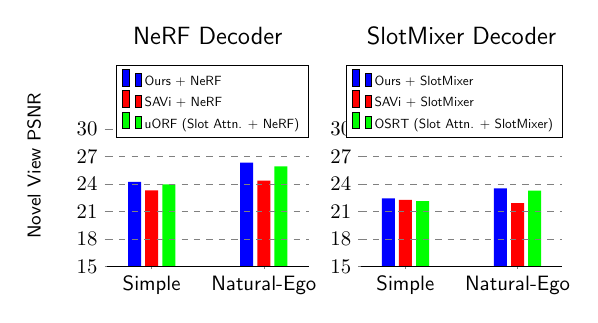}
    \vspace{-4mm}
    \caption{\textbf{Novel View Synthesis in Dynamic 3D Scenes.} We compare the PSNR $(\uparrow)$ of generated views on novel viewpoints. We note improved performance with a large margin relative to SAVi and by a smaller margin relative to static 3D scene models.
    }
    \label{fig:3d-psnr}
\end{figure}

\textbf{Baselines.} We implement our model following the description in Section \ref{sec:appl:3d} and evaluate the proposed encoder paired with two decoders: NeRF \cite{nerfs, uorf, obsurf} and SlotMixer \cite{osrt}. For both decoder choices, we compare the performance by substituting our proposed encoder with: \textit{(i)} SAVi encoder, the current state-of-the-art for learning sequential slot representations from sequential inputs based on recurrence; and \textit{(ii)} Slot Attention applied independently to each time-step. This use of Slot Attention corresponds to the uORF framework of \citet{uorf} when using a NeRF decoder and to the OSRT framework of \citet{osrt} when using a SlotMixer decoder, the two state-of-the-art models for 3D-aware slot learning from static 3D scene observations. Since the focus of this study is on encoder architecture, for fair comparison, we use an identical SLSR backbone and identical decoder architectures across all compared models while substituting only the encoder that produces the slots.

\textbf{Metrics.} To measure performance, we focus on three main aspects of interest: \textit{i)} representation quality, \textit{ii)} novel view synthesis and \textit{iii)} unsupervised segmentation across time and cameras. To measure representation quality, we perform slot linear probing and report the classification accuracy for the categorical object factors and the $R^2$-score for the continuous-valued object factors following \citet{dang2021evaluating} (see Appendix \ref{ax:metric:libprobe}). To measure the performance on novel view synthesis, we report the PSNR on novel views that were not shown to the encoder. For segmentation, we report FG-ARI computed in 4 ways to measure representational consistency: \textit{(i)} per camera per time-step, \textit{(ii)} per camera across time-steps, \textit{(iii)} across cameras per time-step, and \textit{(iv)} across cameras and time-steps.

\subsubsection{Results}

\textbf{3D-Aware Object-Centric Scene Representation and Decomposition.} Fig.~\ref{fig:3d-nerf-probing} and \ref{fig:3d-slotmixer-probing} show our evaluation of linear-probing performance on slots to predict object factors. With both NeRF and SlotMixer decoder, our proposed encoder surpasses the SAVi baseline as well as the static 3D scene models, with a greater gap when the NeRF decoder is used. 
% Additionally, our proposed encoder also outperform the static 3D scene models uORF and OSRT \cite{osrt, uorf, obsurf}. 
A drawback of applying static scene models e.g., uORF and OSRT, on dynamic scenes is that they can suffer in producing aligned slot representations across time, evidenced by the FG-ARI score (reported in Fig.~\ref{fig:ari-3d} in Appendix) computed for measuring temporal consistency.
Another noteworthy point is that while temporal models like SAVi should theoretically benefit from the motion information that static scene models cannot access, SAVi's recurrent implementation and the resulting training instability prevent it from leveraging temporal information effectively---leading to its worse performance than even the static models.

\textbf{Novel View Synthesis.} In Fig.~\ref{fig:3d-psnr}, we report the PSNR of generated views on unseen camera viewpoints i.e., viewpoints that were not given to the encoder to infer the slot representation. This metric measures the model's understanding of the underlying scene geometry as well as the slot representation quality. We observe that our model consistently surpasses both SAVi and per-timestep Slot Attention in performance.

\subsection{Ablations}

\textbf{Learned vs. Random Slot Initialization.} In Fig.~\ref{fig:ablation-2d} and \ref{fig:ablation-3d} in the appendix, we evaluate the impact of initializing slots by randomly sampling them from a learned Gaussian. In this variant, we note a generally worse performance compared to initializing slots as learned parameters.

\textbf{Decoupled vs. Joint Slot Interaction.} In Fig.~\ref{fig:ablation-2d} and \ref{fig:ablation-3d} in the appendix, we also analyze the effect of decoupling the slot-slot self-attention along the time and object axes versus letting all $NT$ slots interact via a single self-attention step. In terms of performance alone, we do not notice a clear advantage of either version. However, it is also important to acknowledge the memory complexity: the joint interaction version requires $\cO(N^2T^2)$ memory 
which can be costlier and hurt scalability compared to using the decoupled version which requires a lower $\cO(NT^2) + \cO(N^2T)$ memory. 

\textbf{No Inverted Attention and Renormalization.} In Fig.~\ref{fig:ablation-2d} in the appendix, we evaluate the impact of using standard dot-product attention instead of inverted attention and renormalization that introduces competition among slots in the bottom-up attention step \cite{slotattention, invertedattn}. Without inverted attention, we find that the video decomposition performance as measured by FG-ARI becomes worse, suggesting that inverted attention is a beneficial inductive bias to keep.

\textbf{Static and Dynamic Field Decoupling in NeRF Decoder.} In Fig.~\ref{fig:ablation-3d-nerf-varieties} in the appendix, we compare the effect of having separate field models for the static topography and the dynamic objects of the 3D scene. We find that the static-dynamic decoupling leads to improved representation quality and object-wise decomposition of the dynamic objects. The benefit of such decoupling is more pronounced in the case of CLEVR-Natural-Ego dataset which is more visually complex than the CLEVR-Simple dataset. 

\section{Discussion}

In conclusion, this work introduces a novel temporally-parallelizable object-centric binding architecture for efficiently processing sequential data to learn slot representations, overcoming the drawbacks of conventional RNN-based architectures. We present two novel auto-encoder models utilizing this architecture for unsupervised object-centric learning, tailored for 2D unposed videos and dynamic 3D scene videos. 
Our comprehensive evaluations across various decoders demonstrate the proposed architecture's superior performance and computational efficiency. 

\textbf{Limitations and Avenues.} We note the limitations of the current work and avenues to address them via future work. First, while our proposed design replaces RNNs with attention, thus providing the benefit of speed and parallelization, it also incurs a quadratic memory complexity in terms of sequence length. To address this, future explorations can consider the use of recent advances e.g., parallelizable SSMs \cite{s4}, based on our framework. Second, it is of interest to apply the proposed architecture to longer episodes and real scenes to utilize the full potential of our scalable approach. Third, the proposed framework can be used to build object-centric dynamics models that can find applications in various agent learning scenarios, e.g., planning for autonomous vehicles, robotics, and RL.

\section*{Impact Statement}
This paper presents a representation learning method for images and videos. While the proposed model uses generation as an objective function, the goal of the proposed model is not on generation but the quality of representation. Therefore, the negative impact of the proposed model on potential fake image generation is little. As the proposed method is very general representation method, it can be used in any application that might include the use of any intention.

\section*{Acknowledgements}
Sungjin Ahn is supported by Brain Pool Plus Program (No. 2021H1D3A2A03103645) through the National Research Foundation of Korea (NRF) funded by the Ministry of Science and ICT. 

\bibliography{refs}
\bibliographystyle{template}

\clearpage
\appendix
\onecolumn
\section{Additional Related Work}
\label{ax:addl_related_work}

\textbf{Additional Works in Object-Centric Learning.} Some works leverage GANs to generate scenes from object-centric noise latents \cite{van2020investigating, blockgan, giraffe}, however, these focus on static scenes and at the same time, do not provide an encoder to learn slots, although such an encoder may be learned as a post-processing step after the GAN is trained. \cite{Du2021UnsupervisedDO} seeks to learn 3D bounding boxes from 2D videos, however, it requires heavy use of manual hand-crafted priors to perform the back-projection from 2D masks to 3D bounding boxes. 

\textbf{3D Scene Representation Learning.} Several works develop auto-encoders for learning to represent 3D scenes \cite{gqn, snp, rsnp, srt, dyst}, including without available camera pose information \cite{sajjadi2023rust}. Another line of work seeks to develop generalizable NeRFs that can learn a NeRF representation without needing per-scene optimization. For this, they are equipped with an encoder that provides a representation given the input views via a single feedforward pass \cite{nerfvae}. However, these representations are not decomposed in an object-centric manner. Static-dynamic decomposition has been pursued in this line \cite{see3d}, however, it does not provide per-object decomposition without post-processing. Another parallel line of work aims to learn implicit scene representations via per-scene NeRF training \cite{unisim, streetsurf, suds, emernerf, wu2022d2nerf}. However, lacking an encoder, expensive training is needed to train them, and representations of distinct scenes carry different semantics. Furthermore, object-centric decomposition requires non-trivial post-processing.

\textbf{Related Architectures for Compression and Attention.} Another line of work seeks to summarize or compress the input activations into a small number of latents without asking for object-centric decomposition of the inputs. If such is the goal, our architecture may still find use in such applications. Along this line, several architectures have been proposed \cite{perceiver, lbanp, rsnp, settransformer, invertedattn}. In implementing the self-attention among slots, our axis-wise attention (along time and object axis) can be seen to be similar to axial attention \cite{axialattention}, which provides memory cost savings.

\section{Additional Experiment Results}
\subsection{FG-ARI Performance on Dynamic 3D Scenes} 
We report and compare the FG-ARI computed along several axes in order to measure segmentation consistency within a view, segmentation across time-steps but with the same camera, segmentation across cameras within a time-step, and segmentation across both cameras and time-steps. We note that when it comes to being consistent across both time-steps and cameras, our encoder outperforms all baselines. We also note that applying static 3D scene models (i.e., OSRT and uORF) per time-step independently can lead to poor consistency in the slots across time, as the slots of each time-step are not aware of the slots of the other time-steps.
\begin{figure}[h]
    \centering
    \begin{subfigure}[b]{0.48\columnwidth}
            \includegraphics[width=\columnwidth]{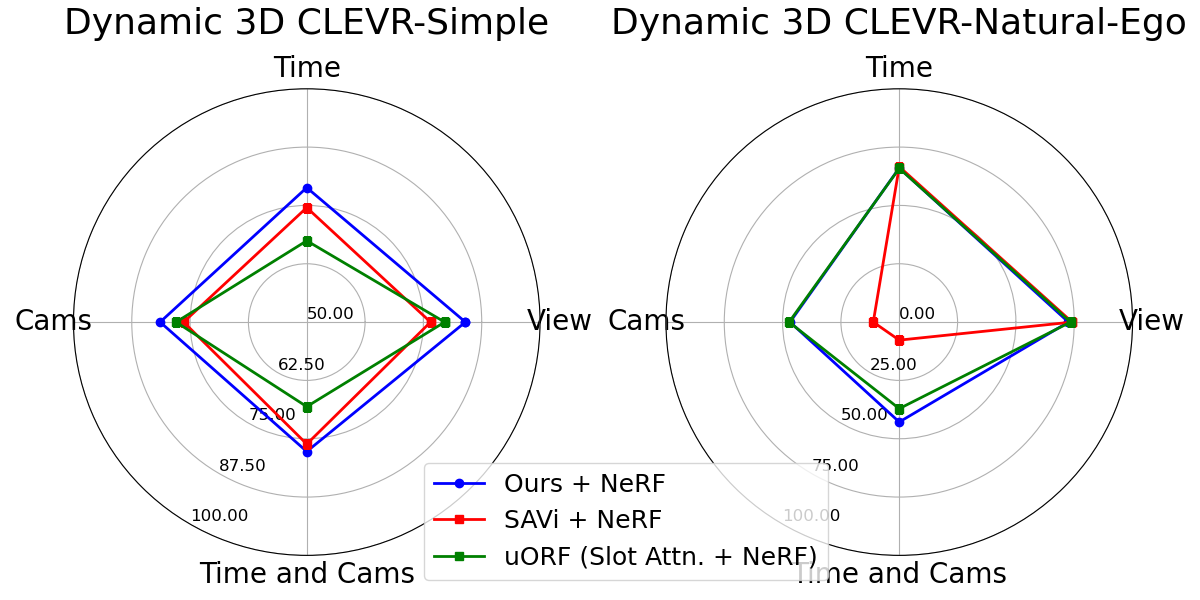}
            \caption{NeRF Decoder}
    \end{subfigure}
    \begin{subfigure}[b]{0.48\columnwidth}
            \includegraphics[width=\columnwidth]{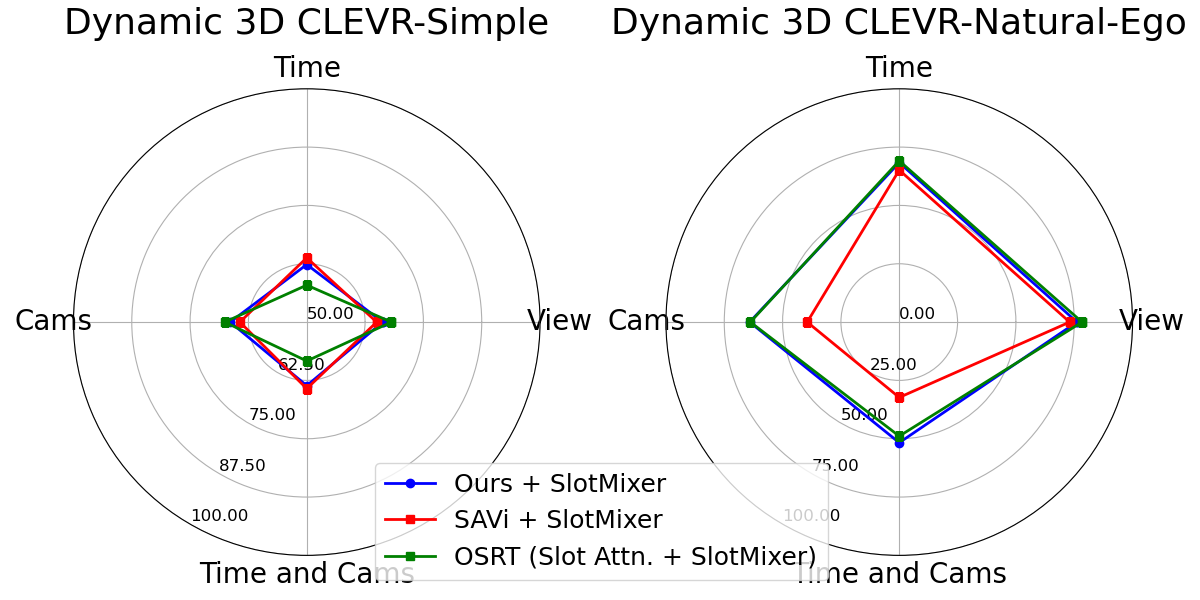}
            \caption{SlotMixer Decoder}
    \end{subfigure}
    \caption{\textbf{FG-ARI Performance on Dynamic 3D Scenes.} We report and compare the FG-ARI computed along several axes in order to measure segmentation consistency within a view, segmentation across time-steps but with the same camera, segmentation across cameras within a time-step, and segmentation across both cameras and time-steps.}
    \label{fig:ari-3d}
\end{figure}

\subsection{Analysis of the Proposed Encoder on Dynamic 3D Datasets} 
\begin{figure}[t]
    \centering
    \begin{subfigure}[b]{0.49\columnwidth}
    \includegraphics[width=\columnwidth]{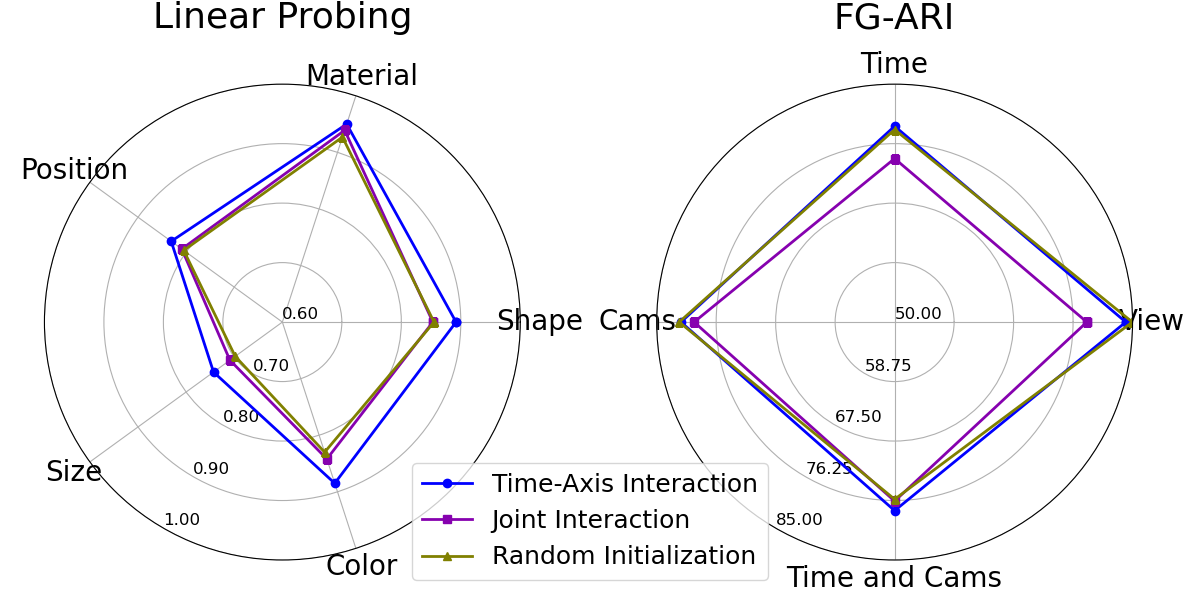}
    \caption{Dynamic 3D CLEVR-Simple}
    \vspace{2mm}
    \end{subfigure}
    \begin{subfigure}[b]{0.49\columnwidth}
    \includegraphics[width=\columnwidth]{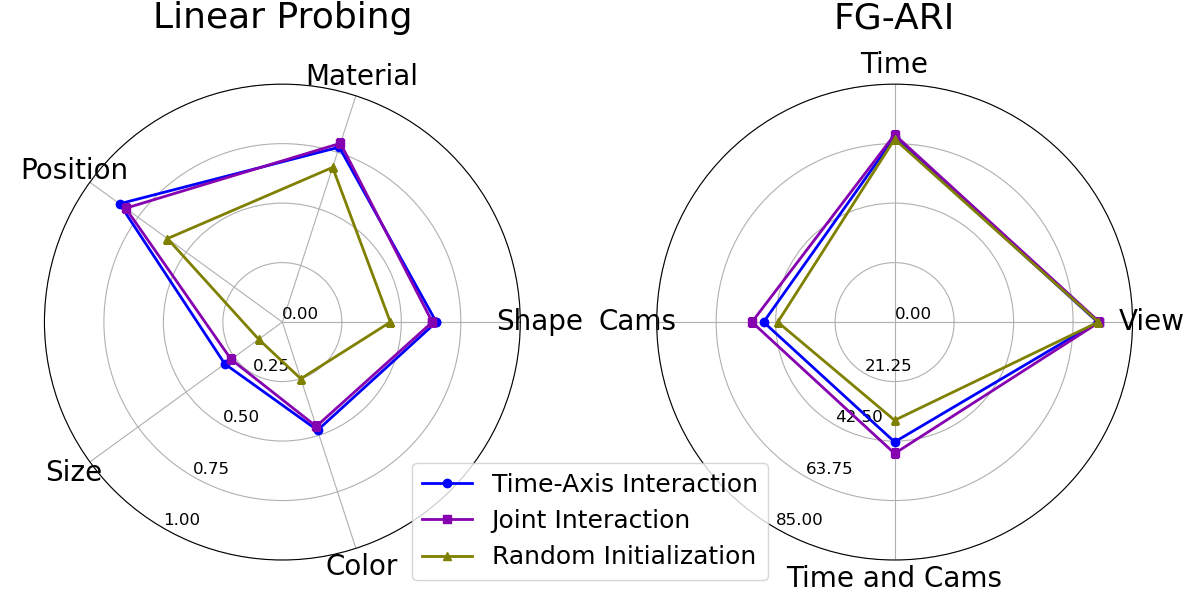}
    \caption{Dynamic 3D CLEVR-Natural-Ego}
    \end{subfigure}
    \vspace{-4mm}
    \caption{\textbf{Analysis of the Proposed Encoder on Dynamic 3D Scene Understanding.} We report and compare the following: \textit{Left:}  Linear-probing performance. \textit{Right:}  Unsupervised segmentation performance. }
    \label{fig:ablation-3d}
\end{figure}

In Fig.~\ref{fig:ablation-3d}, we ablate the design choices underlying our proposed encoder on the dynamic 3D datasets.

\begin{figure}[t]
    \centering
    \begin{subfigure}[b]{0.49\columnwidth}
    \includegraphics[width=\columnwidth]{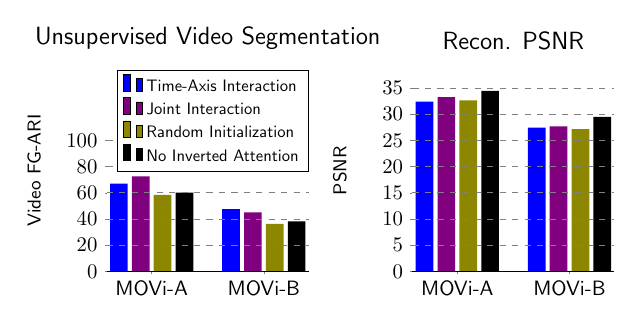}
    \caption{FG-ARI and PSNR}
    \vspace{2mm}
    \end{subfigure}
    \begin{subfigure}[b]{0.49\columnwidth}
    \includegraphics[width=\columnwidth]{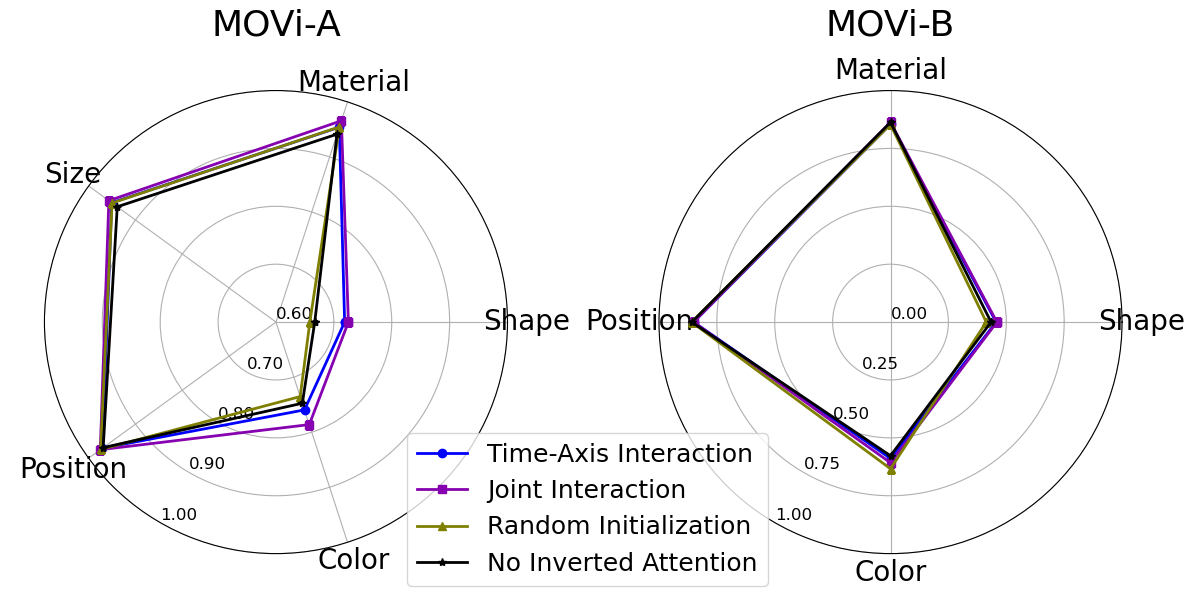}
    \caption{Linear Probes}
    \end{subfigure}
    \vspace{-4mm}
    \caption{\textbf{Analysis of the Proposed Encoder on MOVi-A and MOVi-B.} We report and compare the following: \textit{Left:} Unsupervised segmentation performance and Reconstruction quality. \textit{Right:} Linear-probing performance.}
    \label{fig:ablation-2d}
\end{figure}

\subsection{Effect of Decoupling Static and Dynamic Fields in NeRF Decoder} 
\begin{figure}[h]
    \centering
    \begin{subfigure}[b]{0.48\columnwidth}
            \includegraphics[width=\columnwidth]{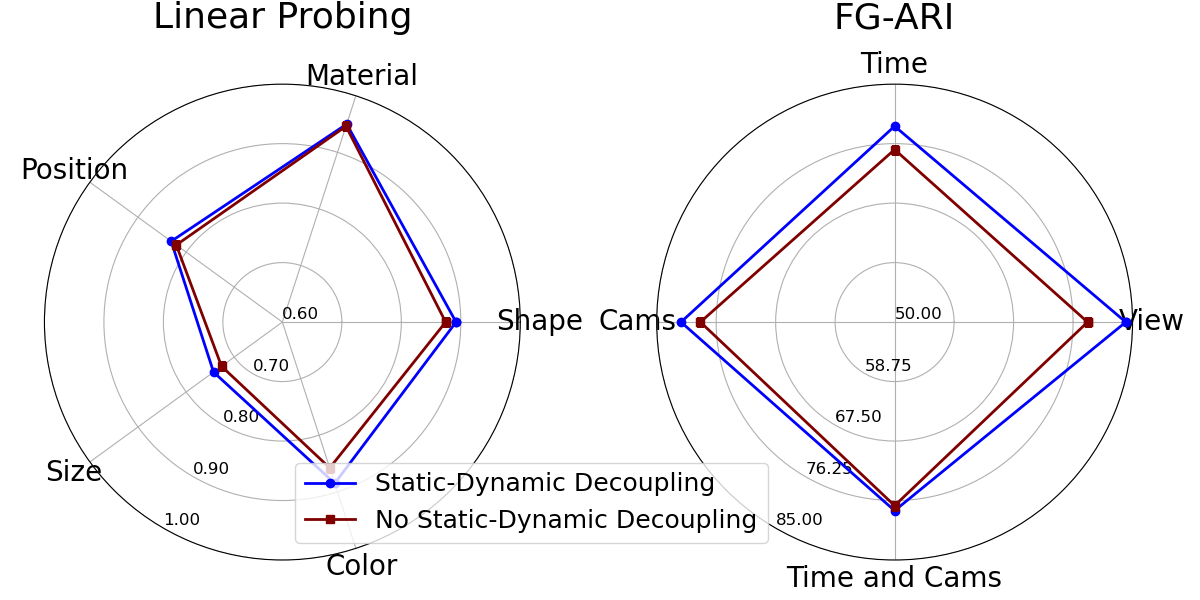}
            \caption{Dynamic 3D CLEVR-Simple}
    \end{subfigure}
    \begin{subfigure}[b]{0.48\columnwidth}
            \includegraphics[width=\columnwidth]{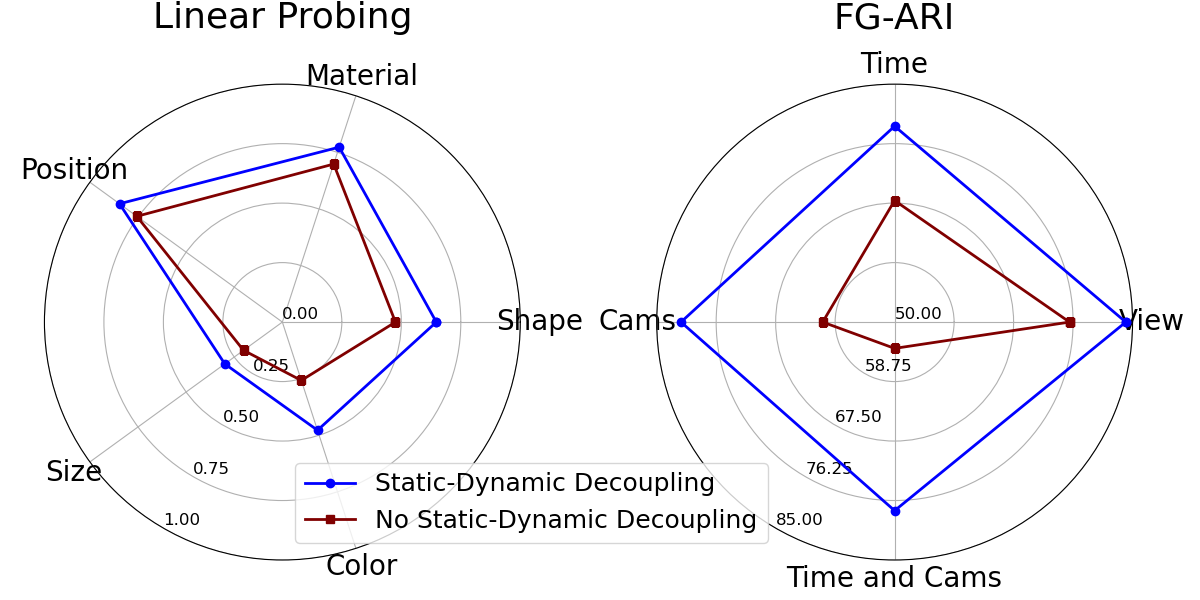}
            \caption{Dynamic 3D CLEVR-Natural-Egocentric}
    \end{subfigure}
    \caption{\textbf{Effect of Static-Dynamic Decoupling in the NeRF Decoder.} We report and compare the following: \textit{Left:} Unsupervised segmentation performance. \textit{Right:} Linear-probing performance.}
    \label{fig:ablation-3d-nerf-varieties}
\end{figure}

In Fig.~\ref{fig:ablation-3d-nerf-varieties}, we ablate this aspect and report the results. We find that static-dynamic decoupling improves performance both in terms of representation quality as suggested by the linear-probing result as well as in terms of unsupervised segmentation. The improvement is more marked in the visually complex and egocentric CLEVR-Natural-Ego dataset.

\subsection{Length Generalization on Longer-Than-Training Sequences}
\begin{figure}[h]
    \centering
    \includegraphics[width=0.5\textwidth]{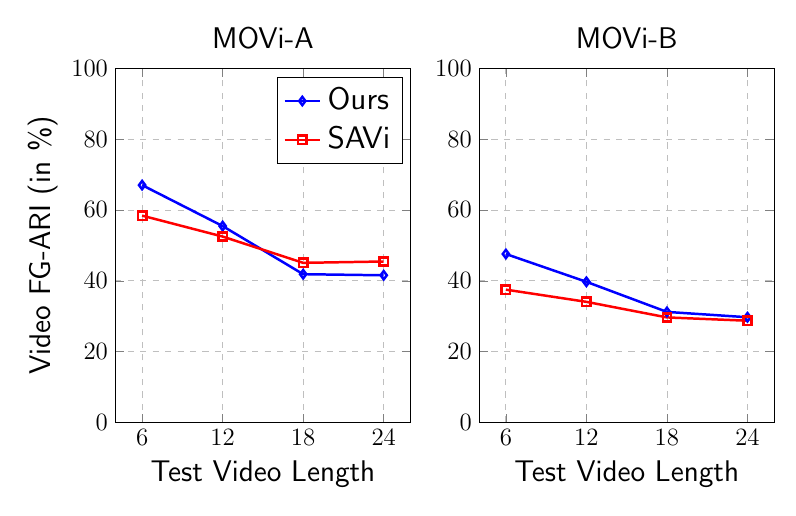}
    \caption{\textbf{Generalization on Sequence Length Longer than Training.} We report the video-level FG-ARI score on MOVi-A and MOVi-B for sequence lengths: 6, 12, 18, and 24; while the training sequence length used is 6. For our model, we use a sliding-window approach to apply the model on longer sequence i.e., the slots of each time-step are inferred from the sequence of most recent 6 frames. For SAVi, we follow the default approach by running the recurrence on the entire available sequence.}
    \label{fig:length-generalization}
\end{figure}

In Fig.~\ref{fig:length-generalization}, we report the video-level FG-ARI score on MOVi-A and MOVi-B for sequence lengths: 6, 12, 18, and 24; while the training sequence length used is 6. For our model, we use a \textit{sliding-window approach} to apply the model on longer sequence i.e., the slots of each time-step are inferred from the sequence of most recent 6 frames. For SAVi, we follow the default approach by running the recurrence on the entire available sequence. On moderately longer sequence lengths i.e., 6 and 12, our model outperforms the baseline SAVi, showing better object-centric decomposition and consistency. On even longer sequences, i.e., sequence lengths 18 and 24, our performance is matched or slightly worse than SAVi, pointing towards an area of improvement for the proposed model. 

\section{Details of Dynamic 3D Scene Datasets}
\label{ax:datasets}
\begin{figure}[h]
    \centering
    \includegraphics[width=1.0\textwidth]{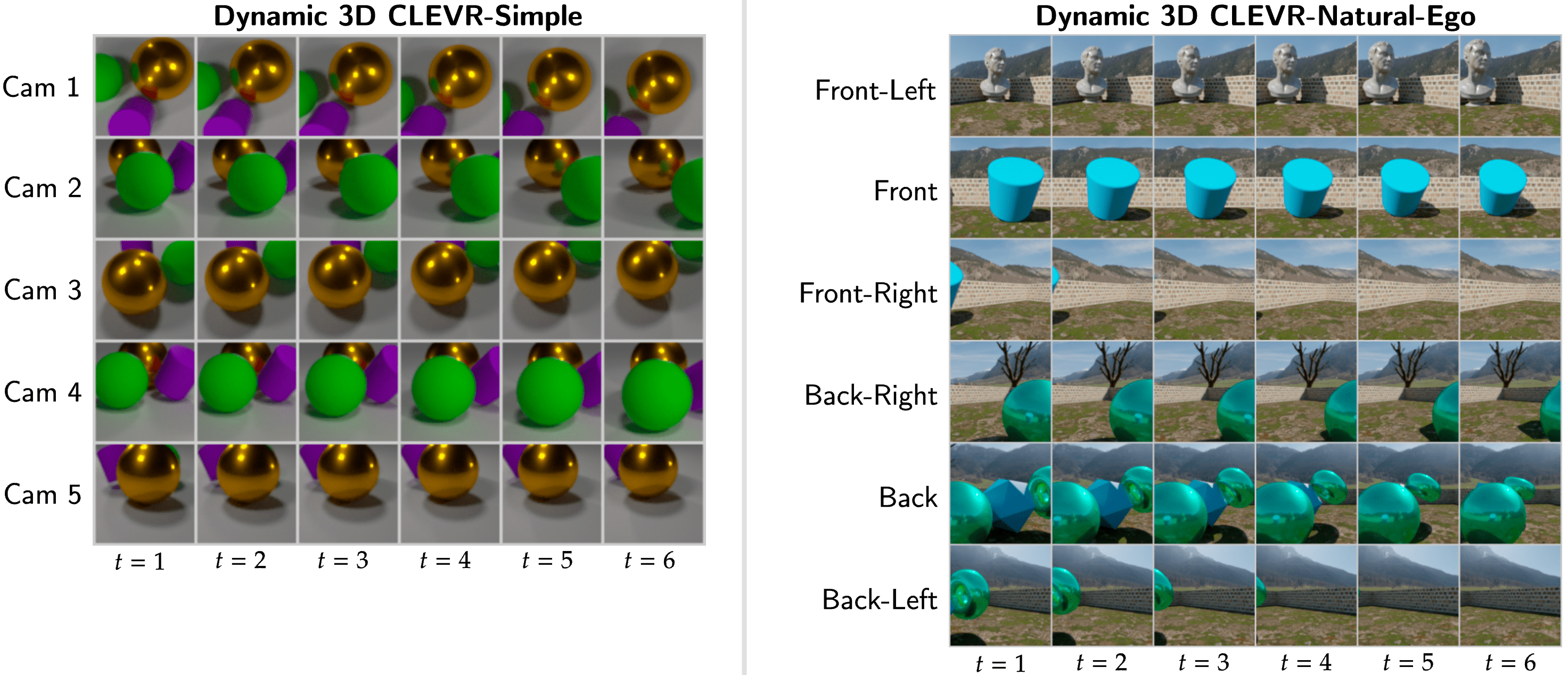}
    \caption{\textbf{Samples of the Proposed Dynamic 3D Scene Datasets.} On the left, we show 5 cameras and 6 time-steps of the Dynamic 3D CLEVR-Simple dataset. On the right, we show 6 ego-centric cameras and 6 time-steps of the Dynamic 3D CLEVR-Natural-Ego dataset.}
    \label{fig:data-samples}
\end{figure}

In this work, we synthesize two dynamic 3D scene datasets to evaluate scene understanding. We provide sample episodes in Fig.~\ref{fig:data-samples}. In the following sections, we provide detailed specifications of these datasets. 
% All datasets and the codebase used to generate them will be publicly released.

\subsection{Dynamic 3D CLEVR-Simple}
The dataset consists of 2000 scene episodes. In each scene episode, the 2-4 objects are randomly instantiated with random velocities heading towards the center of the arena. As the scene plays out, videos are recorded by 5 pin-hole cameras randomly positioned on a hemisphere looking toward the origin of the scene. The objects can take one of four possible shapes: \textit{sphere}, \textit{cylinder}, \textit{cube}, and \textit{monkey}; one of 32 possible colors; one of 2 possible materials: \textit{rubber} and \textit{metal}; and one of 3 possible sizes. The full video length per episode is 32. We train the models on 6-length episodes randomly cropped from the 32-length episodes. The dataset is generated using Blender\footnote{\url{https://www.blender.org}}. We extended the codebase of the original CLEVR dataset\footnote{\url{https://github.com/facebookresearch/clevr-dataset-gen}} to generate this dataset. 

\subsection{Dynamic 3D CLEVR-Nature-Ego}
\begin{figure}[h]
    \centering
    \includegraphics[width=0.5\textwidth]{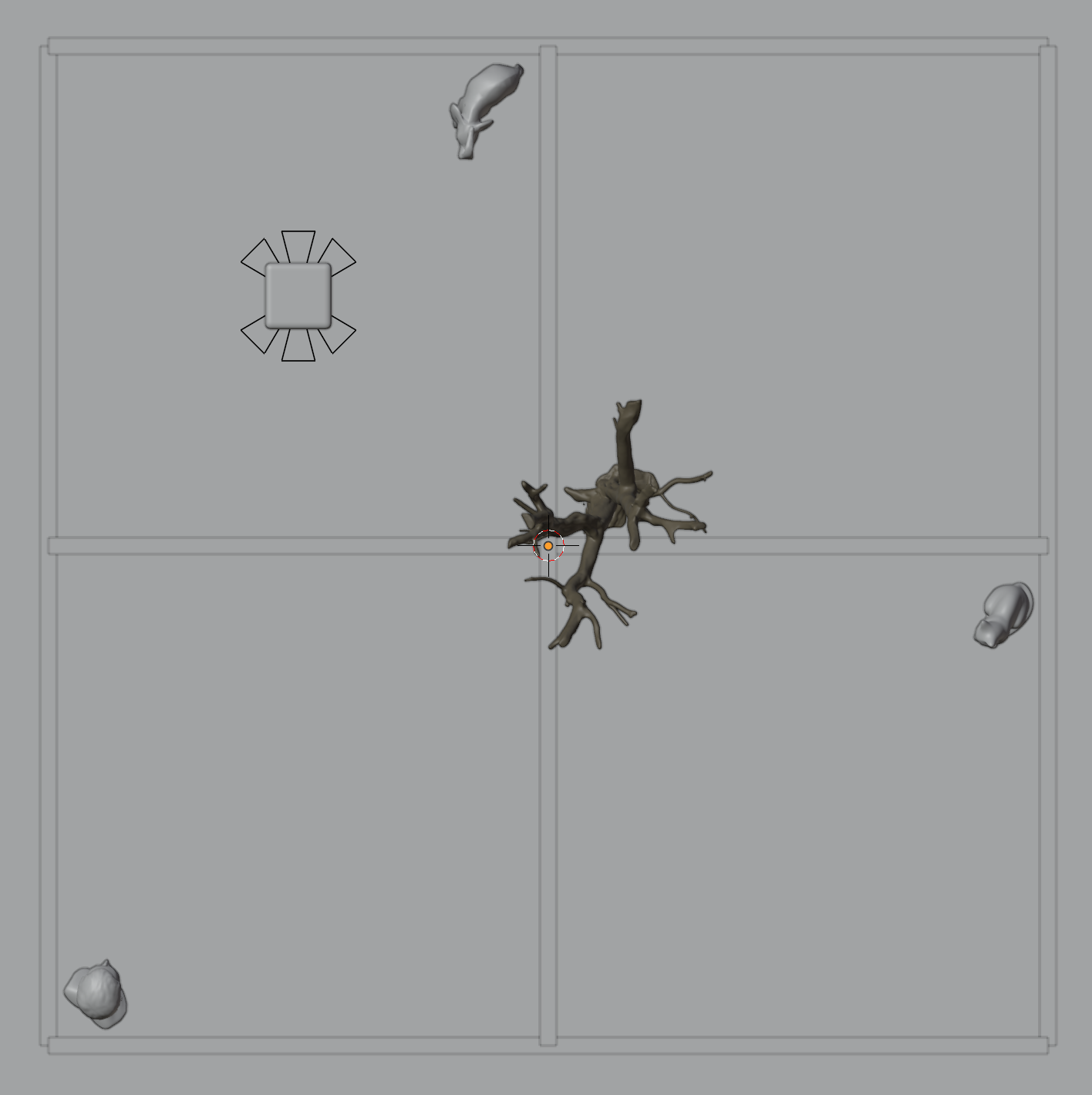}
    \caption{\textbf{Environment Map for CLEVR-Nature-Ego Dataset.} We illustrate the top-view of the static topography and a randomly placed ego object that carries 6 cameras: \texttt{Front\_Left}, \texttt{Front}, \texttt{Front\_Right}, \texttt{Back\_Left}, \texttt{Back}, and \texttt{Back\_Right}. The ego and between 3 to 5 randomly chosen object shapes are spawned inside one of the randomly picked rooms. All objects including the ego are instantiated with a random position and velocity at the start of the episode. Then, the objects and the ego evolve through time under the influence of Blender's physics engine. 24-length RGB video recordings from the 6 ego cameras along with their respective camera poses constitute an episode of the dataset.}
    \label{fig:clevr-ego-static-map}
\end{figure}

The dataset consists of 2000 scene episodes. In each scene episode, the 3-5 objects are randomly instantiated with random heading velocities. An ego object is also instantiated with a random velocity. The ego object is mounted with 6 cameras. As the scene plays out, the objects can collide with each other, with the static topography, and with the moving ego (which can conversely affect the motion of the ego object). The videos are recorded by pin-hole cameras. The objects can take one of six possible shapes: \textit{sphere}, \textit{cylinder}, \textit{cube}, and \textit{cone}, \textit{icosahedron}, and \textit{torus}; one of 32 possible colors; one of 2 possible materials: \textit{rubber} and \textit{metal}; and one of 3 possible sizes. The full video length per episode is 24. We train the models on 6-length episodes randomly cropped from the 24-length episodes. The static topography consists of 4 rooms as visualized in Fig.~\ref{fig:clevr-ego-static-map} and the objects and ego are instantiated in one of the 4 rooms randomly within each episode. Each room consists of a unique static object such as an elephant, a tree, a cat statue, or a statue of a human face. The scene uses realistic lighting, background, and materials. The dataset is generated using Blender\footnote{\url{https://www.blender.org}}. We extended the codebase of the original CLEVR dataset to generate this dataset. 

\section{Additional Model Details}

\subsection{Parallelizable Spatiotemporal Binder}

\subsubsection{Slot Initialization}
The slots can be initialized either via learning or by sampling them from a learned Gaussian. We describe the initialization for both options.

\textbf{Learned Initialization.} We maintain $N$ learned parameters $\ta^\text{slot\_init}_n \in \eR^{D}$ for $n=1, \ldots, N$. Then, the $NT$ slots are initialized as follows.
\begin{align*}
    \bs_{t,n}^{(0)} = \ta^\text{slot\_init}_{n}.
\end{align*}
That is, the initialization is shared for each $n$-th slot across all time-steps. 

\textbf{Random Initialization.} We maintain learned parameters: mean $\boldsymbol{\ta}^\text{slot\_init\_mean} \in \eR^D$ and standard deviation $\boldsymbol{\ta}^\text{slot\_init\_sigma} \in \eR^D$ that are shared for all slots. With these mean and standard deviation parameters, we sample $N$ slots from a Gaussian and then broadcast the $N$ samples across the $T$ time-steps to initialize the $TN$ slots. 
\begin{align*}
    \hat{\bs}_{n}^{(0)} &\sim \cN(\boldsymbol{\ta}^\text{slot\_init\_mean}, \boldsymbol{\ta}^\text{slot\_init\_sigma}), && \forall n \in 1, \ldots, N,\\
    \bs_{t,n}^{(0)} &= \hat{\bs}_{n}^{(0)}, && \forall t \in 1, \ldots, T.
\end{align*}
Same as the case of learned initialization, the initialization is shared for each $n$-th slot across all time-steps. 

\subsubsection{Bottom-Up Attention}
Here, we describe a 1-headed version of the inverted dot-product attention that we employ to implement bottom-up attention in our proposed architecture.
\begin{align*}
Q &= W_Q(Q) &&\in \eR^{N_Q \times D}, \\
K &= W_K(K) &&\in \eR^{N_{K} \times D}, \\
V &= W_V(V) &&\in \eR^{N_{K} \times D}, \\
A &= \texttt{softmax}\left(\frac{QK^T}{\sqrt{D}} + \boldsymbol{\beta}, \quad\texttt{axis=`queries'}\right) &&\in \eR^{N_Q \times N_K},\\
A_{i,j} &= \dfrac{A_{i,j}}{\sum_{j=1}^{N_{K}} A_{i, j}} && \in \eR,\\
\texttt{Attention}(Q, K, V) &= AV && \in \eR^{N_Q \times D},
\end{align*}
where $Q, K, V$ denote queries, keys, and values, respectively, $W_Q$, $W_K$, and $W_V$ denote the linear projection matrices for the queries, keys and the values, respectively, $\boldsymbol{\beta}$ denotes the relative positional bias matrix and $\texttt{Attention}(Q, K, V)$ denotes the result of the attention. The argument $\texttt{axis}$ denotes which axis of the attention weights is the softmax applied to. In this case, we apply softmax along the axis of queries followed by renormalization across keys in line with the competitive attention proposed by \citet{slotattention}.

\textbf{Multi-Headed Implementation.} In the case of multi-headed attention, there are independent linear projection matrices per head. Furthermore, there is an additional output linear projection matrix $W_O$ which takes a concatenation of the outputs of each head and maps it to an output result. In the multi-headed case, the softmax operation is applied across both queries and heads. Furthermore, in the multi-headed case, we have distinct, independently learned relative position bias matrices per each head.

\textbf{Causal Masking.} If a causal mask is provided, then the cells in the attention matrix $\frac{QK^T}{\sqrt{d_k}} + \boldsymbol{\beta}$ that are meant to be masked are replaced with $-\texttt{inf}$.

\subsubsection{Attention Along Time Axis}
This attention is implemented via standard multi-headed dot-product attention with causal masking and adding a relative position bias to the attention matrix.

\subsubsection{Attention Along Object Axis}
This attention is implemented via standard multi-headed dot-product attention. No causal masking or relative position bias is required in this case since this is simply the attention between the slots of the same time-step.

\subsubsection{Non-Linearity}
The non-linearity is implemented via a 2-layer MLP. We use the GELU non-linearity \cite{gelu} in the intermediate layer of the MLP.

\subsubsection{Residual Connections and Layer-Normalizations}
Prior to all attention layers as well as the MLP, we layer-normalize the inputs to the layers. The output of each layer is added to the value of the main branch before the application of the layer. That is, all operations are performed via residual connections, allowing our architecture to be potentially scaled to a deep stack containing a large number of layers.

\subsubsection{Applying to Sequences Lengths Beyond Training Length}
To apply the trained encoder on sequences longer than those shown in training, one may adopt a sliding window approach.

\subsection{Object-Centric Learning on 2D Unposed Videos}

\subsubsection{CNN Backbone}
The CNN backbone maps each image in a video to a set of features. We adopt the backbone that is also used by \citet{savi} and \citet{steve}. We describe it in Table~\ref{tab:conv}. Like the previous works \cite{savi, steve}, we add 2D positional embeddings to the CNN output, flatten it to form a set of features, apply layer normalization to the output, and feed to a 2-layer MLP with hidden size 192 and output size 192.
\begin{table}[h!]
\centering
\caption{The CNN backbone architecture used for object-centric learning on 2D unposed videos.}
\vspace{2mm}
\label{tab:conv}
\begin{tabular}{@{}cccccc@{}}
\toprule
Layer & Kernel Size & Stride & Padding & Channels & Activation \\ \midrule
Conv  & $5\times5$         & 1      & 2       & 192  & ReLU       \\
Conv  & $5\times5$         & 1      & 2       & 192  & ReLU       \\
Conv  & $5\times5$         & 1      & 2       & 192  & ReLU       \\
Conv  & $5\times5$         & 1      & 2       & 192  & None       \\ \bottomrule
\end{tabular}
\end{table}

\subsection{Parallelizable Spatiotemporal Binder}
For the proposed PSB layer, we employ the configurations listed in the following Table \ref{tab:psb_config_2d}.

\begin{table}[htbp]
\centering
\caption{\textbf{PSB Configuration.} The table describes the hyperparameters used in the implementation of the PSB encoder for object-centric learning on 2D unposed videos.}
\label{tab:psb_config_2d}
\vspace{2mm}
\begin{tabular}{@{}lc@{}}
\toprule
Parameter & Value \\
\midrule
Model Dimension & 192 \\
Number of PSB Layers & 3 \\
Number of Attention Heads &  \\
\quad - Bottom-Up Attention Heads & 1 (for matching capacity with SAVi) \\
\quad - Time-Axis Attention Heads & 4 \\
\quad - Object-Axis Attention Heads & 4 \\
MLP Configuration &  \\
\quad - Number of Layers & 2 \\
\quad - Hidden Dimension & 768 \\
\quad - Output Dimension & 192 \\
\bottomrule
\end{tabular}
\end{table}

\subsubsection{Spatial Broadcast Decoder}
\label{ax:sbd}
We implement the spatial broadcast decoder (or simply SBD) in the same manner as \citet{slotattention} and \citet{savi}. The SBD takes slot representations and spatially broadcasts them across a predefined grid, adds positional encodings, and feeds them through a CNN. The implementation details are summarized in Table~\ref{tab:sbd}.

\begin{table}[h!]
\centering
\caption{Spatial Broadcast Decoder architecture for transforming slot representations into RGB and alpha-mixing logits.}
\vspace{2mm}
\label{tab:sbd}
\begin{tabular}{@{}cccccc@{}}
\toprule
Layer & Kernel Size & Stride & Padding & Channels & Activation \\ \midrule
Slot Normalization & - & - & - & - & - \\
Positional Embedding & - & - & - & - & - \\
ConvTranspose2d & $5\times5$ & 2 & 2 (Output Padding: 1) & 64 & ReLU \\
ConvTranspose2d & $5\times5$ & 2 & 2 (Output Padding: 1) & 64 & ReLU \\
ConvTranspose2d & $5\times5$ & 2 & 2 (Output Padding: 1) & 64 & ReLU \\
ConvTranspose2d & $5\times5$ & 2 & 2 (Output Padding: 1) & 3 + 1 & None \\
\bottomrule
\end{tabular}
\end{table}

The CNN outputs the RGB image and the alpha mixing logits. We pass the logits through a softmax operation across the $N$ slot decodings. The softmax output acts as the mixing weights for the $N$ RGB object images corresponding to the $N$ slots, respectively. We use the same positional encoding approach as used in \citet{slotattention} and \citet{savi}. That is, the 2D coordinate of each feature is linearly projected to an appropriately sized embedding.

\subsubsection{Autoregressive Image Transformer Decoder}
\label{ax:autoreg_tf}
For the autoregressive image transformer decoder, we adopt the implementation of SLATE \cite{slate, steve}. That is, conditioned on the slots, we predict the VQ code representation of the image in an autoregressive manner. The VQ code is obtained via a Discrete VAE trained jointly with the rest of the model with codebook size 4096 as prescribed by \citet{slate, steve}. The transformer implementation uses 8 layers with 4-heads and a hidden size of 192.

\subsection{Object-Centric Learning on 3D Posed Multi-Camera Videos}

\subsubsection{Set Latent Scene Representation (SLSR) Backbone}

The Set Latent Scene Representation or SLSR representation is computed by taking multiple posed views of a specific time-step and producing a collection or a set of features or latents, thus called Set Latent Scene Representation \cite{srt, osrt}. This is done by first applying a per-view CNN, flattening the resulting features, stacking these features together for all the views, and lastly, feeding this collection of features to a transformer. 

\textbf{CNN.} In Table \ref{tab:cnn_3d_clevr_simple}, we describe the CNN configuration that we use for the 3D Dynamic \textit{CLEVR-Simple} dataset and in Table \ref{tab:cnn_3d_clevr_natural_ego}. 

\textbf{Transformer.} The specifications of the cross-view transformer are provided in the Table \ref{tab:cross-view-transformer}.

\begin{table}[t]
\centering
\caption{Description of the CNN architecture used to compute SLSR for the dynamic 3D \textit{CLEVR-Simple} dataset.}
\vspace{2mm}
\label{tab:cnn_3d_clevr_simple}
\begin{tabular}{@{}cccccc@{}}
\toprule
Layer & Kernel Size & Stride & Padding & Channels & Activation \\ \midrule
GroupNorm & - & - & - & 192 & - \\ \midrule
\multicolumn{6}{l}{\textbf{Residual Block 1}} \\ 
\qquad GroupNorm & - & - & - & 192 & - \\
\qquad Conv2d & $5\times5$ & 1 & 2 & 192 & ReLU \\
\qquad Conv2d & $5\times5$ & 1 & 2 & 192 & - \\ \midrule
\multicolumn{6}{l}{\textbf{Residual Block 2}} \\
\qquad GroupNorm & - & - & - & 192 & - \\
\qquad Conv2d & $5\times5$ & 1 & 2 & 192 & ReLU \\
\qquad Conv2d & $5\times5$ & 1 & 2 & 192 & - \\ \midrule
\multicolumn{6}{l}{\textbf{Size-Halving}} \\ 
GroupNorm & - & - & - & 192 & - \\ 
Conv2d & $5\times5$ & 2 & 2 & 192 & - \\ \midrule
\multicolumn{6}{l}{\textbf{Residual Block 3}} \\
\qquad GroupNorm & - & - & - & 192 & - \\
\qquad Conv2d & $5\times5$ & 1 & 2 & 192 & ReLU \\
\qquad Conv2d & $5\times5$ & 1 & 2 & 192 & - \\ \midrule
\multicolumn{6}{l}{\textbf{Residual Block 4}} \\
\qquad GroupNorm & - & - & - & 192 & - \\
\qquad Conv2d & $5\times5$ & 1 & 2 & 192 & ReLU \\
\qquad Conv2d & $5\times5$ & 1 & 2 & 192 & - \\
\bottomrule
\end{tabular}
\end{table}

\begin{table}[t]
\centering
\caption{Description of the CNN architecture used to compute SLSR for the dynamic 3D \textit{CLEVR-Natural-Ego} dataset.}
\vspace{2mm}
\label{tab:cnn_3d_clevr_natural_ego}
\begin{tabular}{@{}cccccc@{}}
\toprule
Layer & Kernel Size & Stride & Padding & Channels & Activation \\ \midrule
GroupNorm & - & - & - & 192 & - \\ \midrule
\multicolumn{6}{l}{\textbf{Residual Block 1}} \\ 
\qquad GroupNorm & - & - & - & 192 & - \\
\qquad Conv2d & $5\times5$ & 1 & 2 & 192 & ReLU \\
\qquad Conv2d & $5\times5$ & 1 & 2 & 192 & - \\ \midrule
\multicolumn{6}{l}{\textbf{Residual Block 2}} \\
\qquad GroupNorm & - & - & - & 192 & - \\
\qquad Conv2d & $5\times5$ & 1 & 2 & 192 & ReLU \\
\qquad Conv2d & $5\times5$ & 1 & 2 & 192 & - \\ \midrule
\multicolumn{6}{l}{\textbf{Size-Halving}} \\ 
GroupNorm & - & - & - & 192 & - \\ 
Conv2d & $5\times5$ & 2 & 2 & 192 & - \\ \midrule
\multicolumn{6}{l}{\textbf{Residual Block 3}} \\
\qquad GroupNorm & - & - & - & 192 & - \\
\qquad Conv2d & $5\times5$ & 1 & 2 & 192 & ReLU \\
\qquad Conv2d & $5\times5$ & 1 & 2 & 192 & - \\ \midrule
\multicolumn{6}{l}{\textbf{Residual Block 4}} \\
\qquad GroupNorm & - & - & - & 192 & - \\
\qquad Conv2d & $5\times5$ & 1 & 2 & 192 & ReLU \\
\qquad Conv2d & $5\times5$ & 1 & 2 & 192 & - \\ \midrule
\multicolumn{6}{l}{\textbf{Size-Halving}} \\ 
GroupNorm & - & - & - & 192 & - \\ 
Conv2d & $5\times5$ & 2 & 2 & 192 & - \\ \midrule
\multicolumn{6}{l}{\textbf{Residual Block 5}} \\
\qquad GroupNorm & - & - & - & 192 & - \\
\qquad Conv2d & $5\times5$ & 1 & 2 & 192 & ReLU \\
\qquad Conv2d & $5\times5$ & 1 & 2 & 192 & - \\ \midrule
\multicolumn{6}{l}{\textbf{Residual Block 6}} \\
\qquad GroupNorm & - & - & - & 192 & - \\
\qquad Conv2d & $5\times5$ & 1 & 2 & 192 & ReLU \\
\qquad Conv2d & $5\times5$ & 1 & 2 & 192 & - \\
\bottomrule
\end{tabular}
\end{table}

\begin{table}[t]
\centering
\begin{tabular}{@{}lc@{}}
\toprule
Parameter & Value \\ \midrule
Number of Layers & 3 \\
Number of Heads & 4 \\
Hidden Dimensions & 192 \\ \bottomrule
\end{tabular}
\caption{Cross-View Transformer Specifications.}
\label{tab:cross-view-transformer}
\end{table}

\subsubsection{Parallelizable Spatiotemporal Binder}
For the proposed PSB layer, we employ the configurations listed in the following Table \ref{tab:psb_config_3d}.

\begin{table}[htbp]
\centering
\caption{\textbf{PSB Configuration.} The table describes the hyperparameters used in the implementation of the PSB encoder for object-centric learning on dynamic 3D posed multi-camera videos.}
\label{tab:psb_config_3d}
\vspace{2mm}
\begin{tabular}{@{}lc@{}}
\toprule
Parameter & Value \\
\midrule
Model Dimension & 192 \\
Number of PSB Layers & 3 \\
Number of Attention Heads &  \\
\quad - Bottom-Up Attention Heads & 4\\
\quad - Time-Axis Attention Heads & 4 \\
\quad - Object-Axis Attention Heads & 4 \\
MLP Configuration &  \\
\quad - Number of Layers & 2 \\
\quad - Hidden Dimension & 768 \\
\quad - Output Dimension & 192 \\
\bottomrule
\end{tabular}
\end{table}

\subsubsection{NeRF Decoder}
\label{ax:3ddec:nerf}

The NeRF decoder consists of a feedforward network that takes as input \textit{i)} a slot vector $\bs$, \textit{ii)} 3D coordinate $\bo$, and \textit{iii)} the ray viewing direction $\bd$ and outputs a density $\sigma$ and an RGB color $\bc$. We implement this MLP in the following manner.

\textbf{Encoding Scalars.} Before providing the 3D coordinate of the point and the 3D vector that denotes the viewing direction to an MLP, we would like to encode it as a vector. For this, we use sine-cosine embedding computed in the following manner.
\begin{align*}
    \text{vectorize}(s) = \begin{bmatrix}
\sin(\gamma_1 s) \\
\cos(\gamma_1 s) \\
\sin(\gamma_2 s) \\
\cos(\gamma_2 s) \\
\vdots \\
\sin(\gamma_{D/2} s) \\
\cos(\gamma_{D/2} s)
\end{bmatrix}
\end{align*}
where \( \gamma_i = \pi \cdot 2^{i-1} / \texttt{max\_value} \) for \( i = 1, 2, \ldots, D/2 \), \(D\) is the dimensionality of the output vector, and \(\texttt{max\_value}\) is a predefined maximum value that the scalar may take. In this work, we choose $D=16$.

\textbf{Network.} The network consists of two MLP blocks. The first MLP block takes the embedding of $\bo$ and the slot vector $\bs$ and outputs an intermediate representation. The MLP is specified in Table \ref{tab:nerf_mlp_block_a}. This intermediate representation is mapped to the density value $\sigma$ via a linear head. Next, the same intermediate representation is added to an embedding of the ray direction $\bd$ and is given to a second MLP block to output another hidden representation. This hidden representation is mapped to the color value $\bc$ via a second linear head. This second MLP block is described in Table \ref{tab:nerf_mlp_block_b}.

\textbf{Static-Dynamic Decoupling and the Sky Head.} To decouple the static and dynamic fields, we learn a dedicated NeRF decoder with separate weights that do not consume slots as input. The idea is to allow it to capture the background static topography that remains fixed across all episodes of the dataset without conditioning on slots whose role is to capture the dynamic and variable elements of the scene. To implement the static field, we learn two separate decoders, one for the static topography and another for the sky or the far field. For capturing static topography, the decoder implementation is identical to that described above except that no slot input is given to the network. For implementing the sky field, we take an MLP architecture identical to the first MLP block of the object field implementation described above (and in Table \ref{tab:nerf_mlp_block_a}). To this, we provide an encoding of the ray direction as input (instead of providing an embedding of the 3D point on the ray), take the output of this MLP, and map it to a color value via a linear head. To obtain the color of an image pixel incorporating the sky head, we shoot the corresponding ray from the camera, sample $N_\text{bins}$ points along the ray, and integrate the colors along the ray as:
\begin{align*}
    \sum_{i=1}^{N_\text{bins}} T_i \alpha_i \bc_i + T_{N_\text{bins} + 1}\bc_\text{sky},
\end{align*}
where $T_i = \prod_{j=1}^{i-1} (1 - \alpha_i)$ is the transmittance, $\alpha_i = 1 - \exp (-\sigma_i ||\bo_{i+1} - \bo_i||_2) $ is the opacity and $\bc_\text{sky}$ is the sky color obtained from the sky head for the ray's viewing direction. An overview of the NeRF with a separate sky head, static head, and object head is shown in Fig.~\ref{fig:nerf_and_slotmixer} (left).

\textbf{Shadow Head.} Another element of modeling visually complex scenes is to model the scene lighting and shadows. To capture the shadows explicitly within the NeRF framework, we can optionally add a shadow head to obtain a shadow coefficient from the intermediate hidden state output by the first MLP block. The aim of the shadow coefficient $\rho$ is to suppress the color value that would have been had there been no shadow at that 3D position. Specifically, the following formula is used to update the \textit{shadeless} color value of a 3D point.

\begin{align*}
    \sigma = \sigma_\text{static} + \sum_{n=1}^N \sigma_n, && \mathbf{c}_\text{shadeless} = \frac{\sigma_\text{static} \bc_\text{static} + \sum_{n=1}^N \bc_n \sigma_n}{\sigma}, && \mathbf{c} =  \left[\rho_\text{static}\prod_{n=1}^N \rho_n\right]\mathbf{c}_\text{shadeless}
\end{align*}

\begin{table}[h]
\centering
\begin{tabular}{lc}
\toprule
\textbf{Layer} & \textbf{Configuration} \\
\midrule
LayerNorm & 192 \\
\cmidrule(lr){1-2}
\multicolumn{2}{l}{\textbf{Residual Block 1}} \\
\qquad LayerNorm & 192 \\
\qquad Linear & $192 \rightarrow 192$ \\
\qquad ReLU & - \\
\qquad Linear & $192 \rightarrow 192$ \\
\cmidrule(lr){1-2}
\multicolumn{2}{l}{\textbf{Residual Block 2}} \\
\qquad LayerNorm & 192 \\
\qquad Linear & $192 \rightarrow 192$ \\
\qquad ReLU & - \\
\qquad Linear & $192 \rightarrow 192$ \\
\bottomrule
\end{tabular}
\caption{Specifications of the first block of the MLP implementation of the NeRF decoder.}
\label{tab:nerf_mlp_block_a}
\end{table}

\begin{table}[h]
\centering
\begin{tabular}{lc}
\toprule
\textbf{Layer} & \textbf{Configuration} \\
\midrule
LayerNorm & 192 \\
\cmidrule(lr){1-2}
\multicolumn{2}{l}{\textbf{Residual Block 1}} \\
\qquad LayerNorm & 192 \\
\qquad Linear & $192 \rightarrow 192$ \\
\qquad ReLU & - \\
\qquad Linear & $192 \rightarrow 192$ \\
\bottomrule
\end{tabular}
\caption{Specifications of the second block of the MLP implementation of the NeRF decoder.}
\label{tab:nerf_mlp_block_b}
\end{table}

\begin{table}[h]
\centering
\begin{tabular}{lc}
\toprule
\textbf{Layer} & \textbf{Configuration} \\
\midrule
LayerNorm & 192 \\
\cmidrule(lr){1-2}
\multicolumn{2}{l}{\textbf{Residual Block 1}} \\
\qquad LayerNorm & 192 \\
\qquad Linear & $192 \rightarrow 192$ \\
\qquad ReLU & - \\
\qquad Linear & $192 \rightarrow 192$ \\
\cmidrule(lr){1-2}
\multicolumn{2}{l}{\textbf{Residual Block 2}} \\
\qquad LayerNorm & 192 \\
\qquad Linear & $192 \rightarrow 192$ \\
\qquad ReLU & - \\
\qquad Linear & $192 \rightarrow 192$ \\
\cmidrule(lr){1-2}
\multicolumn{2}{l}{\textbf{Residual Block 3}} \\
\qquad LayerNorm & 192 \\
\qquad Linear & $192 \rightarrow 192$ \\
\qquad ReLU & - \\
\qquad Linear & $192 \rightarrow 192$ \\
\cmidrule(lr){1-2}
\multicolumn{2}{l}{\textbf{Residual Block 4}} \\
\qquad LayerNorm & 192 \\
\qquad Linear & $192 \rightarrow 192$ \\
\qquad ReLU & - \\
\qquad Linear & $192 \rightarrow 192$ \\
\bottomrule
\end{tabular}
\caption{MLP Specifications for the SlotMixer decoder.}
\label{tab:slotmixer_mlp}
\end{table}

\subsubsection{SlotMixer Decoder}
\label{ax:3ddec:slotmixer}

\begin{figure}[h]
    \centering
    \hfill
    \includegraphics[width=0.30\columnwidth]{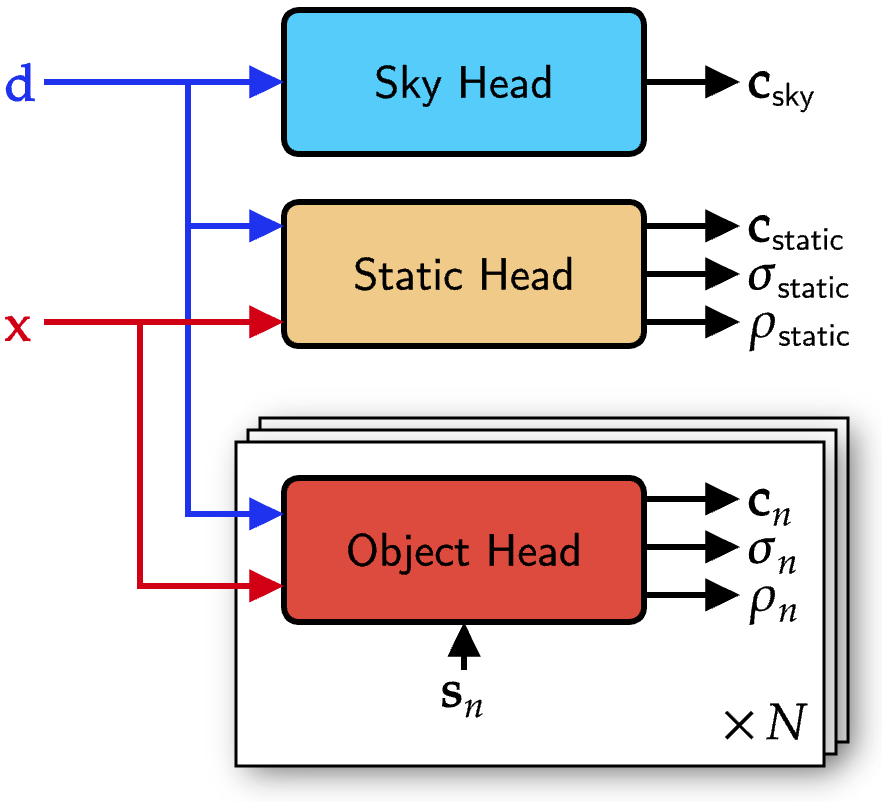} 
    \hfill
    \includegraphics[width=0.48\columnwidth]{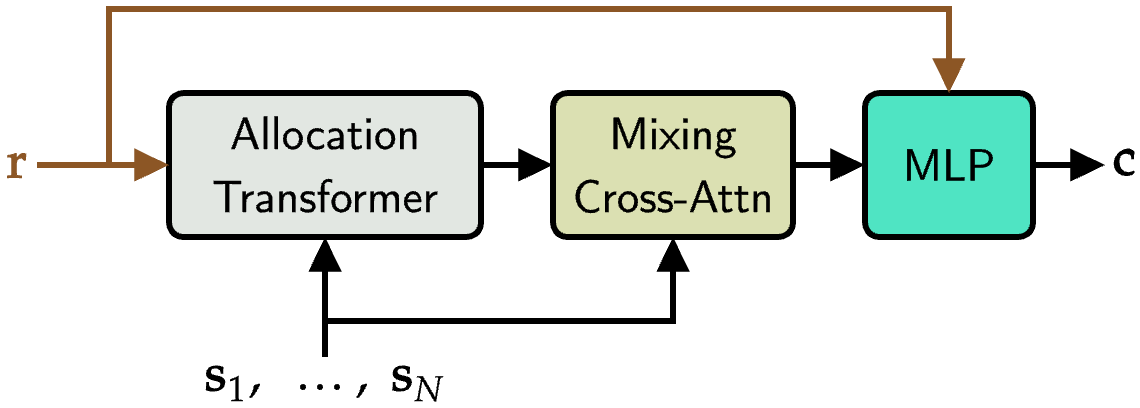}
    \hfill
    \caption{\textbf{3D Decoders.} 
    \textit{Left:} Our NeRF decoder is parametrized in terms of three MLPs to model the sky, the static topography and the objects, respectively. \textit{Right:} SlotMixer decoder \cite{osrt} directly maps a camera ray to a pixel color conditioned on the slots via a transformer-like architecture.}
    \label{fig:nerf_and_slotmixer}
\end{figure}

The SlotMixer decoder directly maps an embedding of the ray shooting from the camera and maps it to a color value via a transformer-like implementation that conditions this function on the slot representation of the scene. Our implementation follows \citet{osrt} whose overview is provided in Fig.~\ref{fig:nerf_and_slotmixer} (right). For the allocation transformer, we use 3 transformer layers with 4-headed attention and hidden size 192. We follow \citet{osrt} in implementing the mixing block as prescribed in the original paper. To implement the MLP, we use the specification provided in Table \ref{tab:slotmixer_mlp}.

\subsection{Optimization}
To train the models, we use a linear learning rate warm-up in the first 30000 training steps and use exponential decay thereafter with a half-life of 1M steps. We use a peak learning rate of $3e-4$. All models are trained to 300K steps. We used a batch size of 24 episodes with 6 time-steps per episode. We use the AdamW optimizer \cite{adamw} with $\beta_1=0.9$ and $\beta_2=0.95$.

\section{Additional Evalutation Details}

\subsection{Permutation-Invariant Slot Linear Probing}
\label{ax:metric:libprobe}

To perform permutation-invariant slot linear probing, we are provided with pre-trained and frozen slot representations of an episode denoted as $\bs_{1:T, 1:N}$. The ground-truth positions of the scene objects of an episode are denoted by $\by^\text{position}_{1:T, 1:M}$, where $M$ represents the number of actual objects in the scene. Let $\bs_{1:T, \pi}$ represent a permutation of $N$ slots taken $M$ at a time, applying the same permutation $\pi$ across all time steps. The first step involves determining the correct permutations $\pi_1, \ldots, \pi_B$ for all episodes within a large batch of $B$ episodes. These permutations are used to correctly assign each slot to an object's label for training the linear probes. To determine these permutations, we follow the methodology outlined in \cite{dang2021evaluating}. We employ an approach akin to the EM algorithm, starting with arbitrary permutations $\pi_1, \ldots, \pi_B$. During the E-step, linear probes are trained using these permutations to predict the ground truth object positions $\by^\text{position}_{1:T, 1:M}$. Subsequently, in the M-step, we iterate over all possible permutations to identify the one that best minimizes the probing error. The E and M steps are repeated until convergence is achieved. Finally, the determined permutations $\pi_1, \ldots, \pi_B$ are used to predict all object factors, such as color $\by^\text{color}_{1:T, 1:M}$, shape $\by^\text{shape}_{1:T, 1:M}$, size $\by^\text{size}_{1:T, 1:M}$, etc. and performance is reported.

\end{document}